\pdfoutput=1

\documentclass[11pt]{article}

\usepackage[preprint]{ACL2025}

\usepackage{times}
\usepackage{latexsym}
\usepackage{tikz}
\usepackage{colortbl}  
\usepackage{tcolorbox}
\tcbuselibrary{breakable}

\usepackage{booktabs}       
\usepackage{amsfonts}       
\usepackage{nicefrac}       
\usepackage{microtype}      
\usepackage{listings}

\usepackage{latexsym}
\usepackage{makecell}
\usepackage{epsfig}
\usepackage{graphicx}
\usepackage{amsmath}
\usepackage{amssymb}
\usepackage{subfig}
\usepackage{booktabs}
\usepackage{multirow}
\usepackage{array}
\usepackage{color}
\usepackage{soul}
\usepackage{transparent}
\usepackage{tabularx}
\usepackage{tabularray}
\UseTblrLibrary{booktabs}
\usepackage{longtable}
\usepackage{algorithmic}
\usepackage[ruled,linesnumbered]{algorithm2e}

\usepackage{url}
\usepackage{subcaption}
\usepackage{enumitem}
\usepackage{booktabs} %
\usepackage{graphicx} %
\usepackage{geometry} %
\usepackage{hyperref}
\definecolor{lightgray}{gray}{0.95}
\definecolor{lightblue}{rgb}{0.87, 0.94, 1}
\definecolor{ADD}{rgb}{1, 0, 0}

\makeatletter
\newcommand\myfont{\@setfontsize\myfont{9.2}{9.5}} 
\newcommand\fontwo{\@setfontsize\fontwo{8.2}{10.5}}
\newcommand\fonthree{\@setfontsize\fontwo{9.2}}
\makeatother

\usepackage[T1]{fontenc}

\usepackage[utf8]{inputenc}

\usepackage{microtype}

\usepackage{inconsolata}

\usepackage{graphicx}

\def \OURS{SciCUEval}

\title{\OURS{}: A Comprehensive Dataset for Evaluating Scientific Context Understanding in Large Language Models}

\author{
  Jing Yu$^{1,2}$\thanks{Equal contribution.}, Yuqi Tang$^{3*}$, Kehua Feng$^{1,4}$, Mingyang Rao$^{1,4}$, Lei Liang$^{5}$, Zhiqiang Zhang$^{5}$, Mengshu Sun$^{5}$,\\ \textbf{Wen Zhang$^{4}$, Qiang Zhang$^{3}$, Keyan Ding$^{1\dag}$, Huajun Chen$^{1,4}$\thanks{Corresponding authors.}} \\
  $^1$ZJU-Hangzhou Global Scientific and Technological Innovation Center, Zhejiang University\\
  $^2$The Polytechnic Institute, Zhejiang University 
  $^3$ZJU-UIUC, Zhejiang University\\
  $^4$College of Computer Science and Technology, Zhejiang University \\
  $^5$AntGroup\\
  \texttt{\{yujing17, dingkeyan, huajunsir\}@zju.edu.cn} \\
}

\begin{document}
\maketitle

\begin{abstract} 
Large Language Models (LLMs) have shown impressive capabilities in contextual understanding and reasoning. However, evaluating their performance across diverse scientific domains remains underexplored, as existing benchmarks primarily focus on general domains and fail to capture the intricate complexity of scientific data.
To bridge this gap, we construct \textbf{\OURS{}}, a comprehensive benchmark dataset tailored to assess the scientific context understanding capability of LLMs. It comprises ten domain-specific sub-datasets spanning biology, chemistry, physics, biomedicine, and materials science, integrating diverse data modalities including structured tables, knowledge graphs, and unstructured texts.
\OURS{} systematically evaluates four core competencies: \emph{Relevant information identification}, \emph{Information-absence detection}, \emph{Multi-source information integration}, and \emph{Context-aware inference}, through a variety of question formats. 
We conduct extensive evaluations of state-of-the-art LLMs on \OURS{}, providing a fine-grained analysis of their strengths and limitations in scientific context understanding, and offering valuable insights for the future development of scientific-domain LLMs.
\end{abstract}

\section{Introduction}

Large Language Models (LLMs) have demonstrated strong capabilities in natural language understanding, reasoning, and generation across a wide range of general-domain tasks \cite{bai2023qwen, openai2024gpt4ocard, dubey2024llama, qwen2025qwen25technicalreport}. However, their application to scientific domains remains challenging due to the unique characteristics of scientific language and knowledge. Scientific texts are often dense with technical terminology, implicit assumptions, multimodal data representations, and tightly interlinked concepts that require deeper contextual comprehension \cite{beltagy2019scibert, mann2020language}.

Existing LLM benchmarks in scientific domains \cite{chen2025benchmarking, sun2024scieval, feng2024sciknoweval, saikh2022scienceqa, pedersen2020sciq, rubungo2024llm4mat, jiang2025benchmarking} primarily focus on direct question-answering tasks, offering limited insight into how well LLMs perform in scientific context understanding, particularly for \textit{noisy} and \textit{lengthy} contexts. Additionally, they often neglect the heterogeneous and structured nature of scientific data, which can span textual descriptions, relational graphs\cite{talmor2018web, he2024g}, and tabular datasets\cite{fang2024large}.  In contrast, robust scientific context understanding demands precise information extraction, the ability to identify gaps or missing elements in context, and the integration of multiple evidence sources to support accurate conclusions.

\begin{figure*}[htbp]
    \centering
    \includegraphics[width=0.98\textwidth]{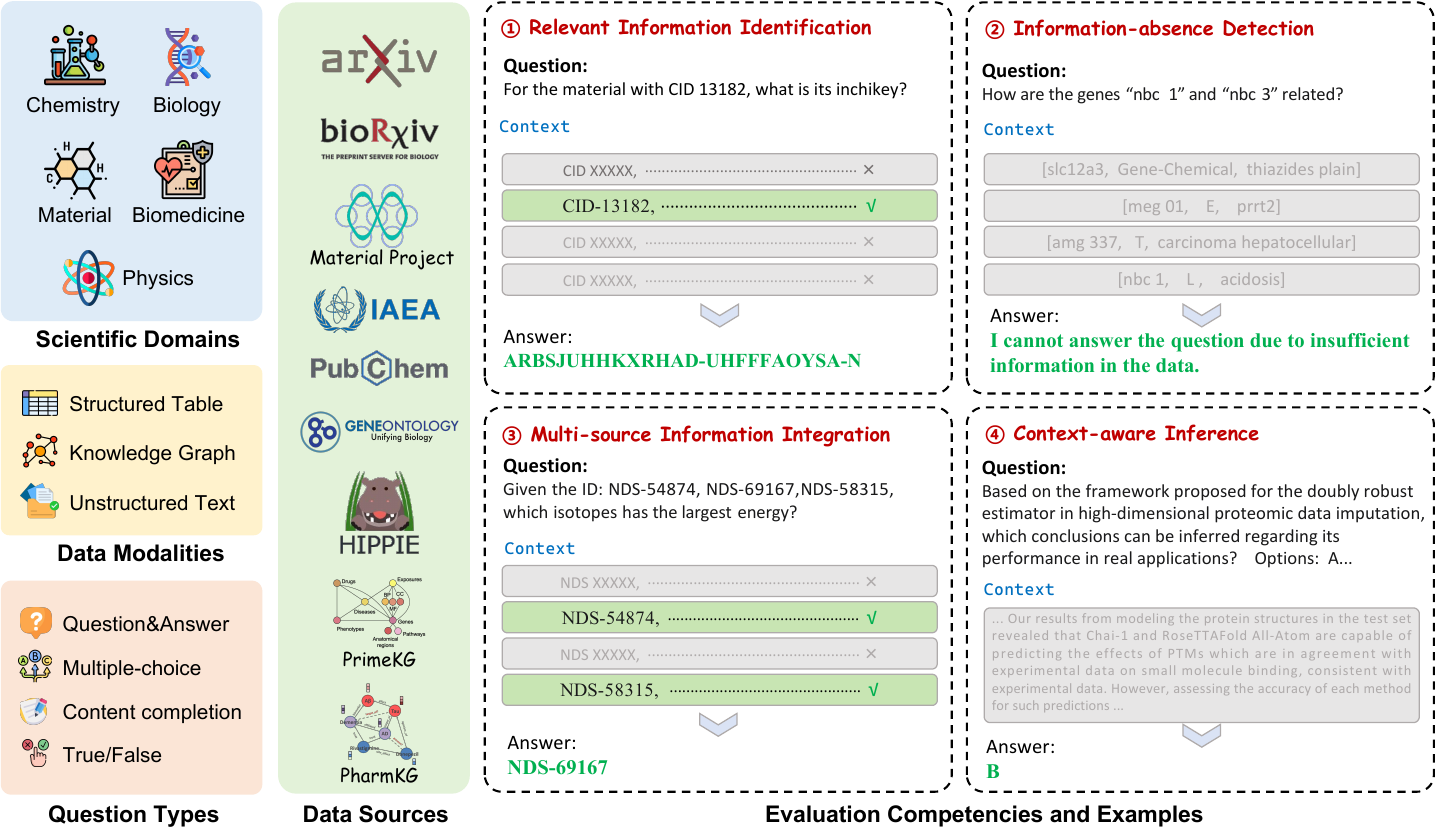} 
    \caption{Overview of the \OURS{} dataset. It spans five scientific domains, supports three data modalities (structured tables, knowledge graphs, and unstructured text), and includes four question types. Data are collected from high-quality scientific sources. The dataset enables evaluation across four key competencies: (1) relevant information identification, (2) information-absence detection, (3) multi-source information integration, and (4) context-aware inference.}
    \label{fig:overview} 
\end{figure*}

To address this gap, we introduce \OURS{}, a comprehensive benchmark dataset designed to rigorously evaluate the scientific context understanding capabilities of LLMs. As shown in Figure~\ref{fig:overview}, \OURS{} spans ten domain-specific sub-datasets covering diverse disciplines such as biology, chemistry, physics, biomedicine, and materials science. Each subset incorporates rich external knowledge in multiple forms (structured tables, semi-structured knowledge graphs, and unstructured scientific texts) to represent the data modalities commonly encountered in scientific research.

\OURS{} targets four core competencies essential for scientific understanding: (1) {Relevant Information Identification} that locate and extract relevant information from complex and lengthy inputs; (2) {Information-Absence Detection} that recognize missing, ambiguous, or incomplete contextual elements; (3) {Multi-source Information Integration} that aggregate and compare information from diverse sources; and (4) {Context-Aware Inference} that deduce accurate conclusions grounded in scientific contexts. These competencies are evaluated using diverse question types, including open-ended Q\&A, multiple-choice, content completion, and true/false validation.

The contributions of this paper are summarized as follows:

\begin{itemize}[left=0pt]
    \item \textbf{Establishing a scientific context understanding benchmark:} We establish a benchmark to evaluate context understanding capabilities of LLMs in scientific domains, serving as a standardized evaluation suite for assessing LLMs' capabilities in identifying, detecting, integrating, and reasoning over scientific contexts.
    \item \textbf{Constructing a diverse set of domain-specific context understanding datasets:} We construct ten sub-datasets across multiple disciplines, encompassing various data modalities and a wide range of question types to ensure comprehensive evaluation.
    \item \textbf{Extensive evaluation and analysis of LLMs:} We systematically evaluate and analyze the performance of various state-of-the-art LLMs on \OURS{}, highlighting their strengths and limitations, and offering insights for improvement. 
\end{itemize}

\begin{table*}[t]
    \small
    \centering
    \caption{Comparison of \OURS{} with existing benchmark datasets. Question Types: QA (Question Answering), MCQ (Multiple Choice Question), T/F (True/False Question), and CC(Cloze Completion).}
    \renewcommand{\arraystretch}{1.3} 
    \resizebox{1\linewidth}{!}{ 
        \begin{tabular}{ccccccc}
            \toprule
            \textbf{Datasets} & \textbf{Contexts} &\textbf{Domains} & \textbf{Data Modalities} & \textbf{Question Types} & \textbf{Evaluation Competencies}  & \textbf{\# Nums} \\
            \midrule
            LongICLBench \cite{li2024long} &  $\checkmark$ &  General &  Text & QA & Identification  & 2,618\\
            LongBench \cite{bai2023longbench} & $\checkmark$  & General, Code  &  Text &  QA &  Identification, Integration &4,750 \\
            LongBench V2 \cite{bai2024longbench} & $\checkmark$  &  General, Law, Finance & Text & MCQ &  Identification, Integration, Inference & 503\\
            RGB\cite{chen2024benchmarking}  & $\checkmark$  & General  &  Text &  QA & Identification, Detec., Integration, Inference   & 1,000\\

            ChemLit-QA \cite{wellawatte2025chemlit} & $\checkmark$ & Chemistry  & Text  &  QA & Identification, Detec., Inference  &1,054 \\
            CHEMRAG-BENCH \cite{zhong2025benchmarking}  & $\times$  & Chemistry  &  Text &  QA, MCQ& Identification, Inference  & 1,932\\
            \midrule
            \textbf{\OURS{}} & $\checkmark$ & Comprehensive Science& Text, Table, KG  &  QA, MCQ, T/F, CC &  Identification, Detec., Integration, Inference & 11,343\\

            \bottomrule
        \end{tabular}
    }
    \label{tab:benchmark_comparison} 
    \vspace{-1em}
\end{table*}

\section{Backgrounds and Related Works}

\paragraph{Context Understanding Tasks}
Large language models (LLMs) have demonstrated remarkable capabilities in context understanding\cite{huang2023advancing, dong2022survey}. This paradigm allows LLMs to adapt flexibly to new external knowledge without requiring additional fine-tuning or retraining\cite{lewis2020retrieval, liu2024repoqa}. Importantly, LLMs have shown strong performance across a wide range of domains when equipped with additional context. Furthermore, in-context learning strengthens LLMs' transparency by firmly establishing their arguments in the documents that were obtained \cite{mialon2023augmented, xiong2024benchmarking}. However, despite its potential, context understanding remains sensitive to prompt design and the quality of the provided context. It also demands that the model possess a robust ability to process and comprehend long texts. In this study, we systematically investigate the robustness and effectiveness of LLMs for context understanding in diverse scientific domains.

\paragraph{Context Understanding Benchmarks} 
 
The development of robust and comprehensive benchmarks for evaluating long context understanding has gained increasing attention in recent research \cite{li2024long, liu2024repoqa, bai2023longbench, zhang2024infty,wu2025tablebench,chen2020hybridqa, chen2021finqa}. For instance, LongICLBench \cite{li2024long} evaluates the long in-context learning capabilities of large language models across various domains, including emotion classification, intent detection, relation extraction, and named entity recognition. RepoQA \cite{liu2024repoqa} evaluates LLMs' long-context code understanding ability. LongBench \cite{bai2023longbench} and LongBench-V2 \cite{bai2024longbench} are comprehensive benchmarks for evaluating long-context understanding and reason across domains including law, finance, literature, news, and code. In the scientific domain, ChemLit-QA\cite{wellawatte2025chemlit} is a chemistry dataset of question-answer-context triplets. CHEMRAG-BENCH\cite{zhong2025benchmarking} benchmarks retrieval-augmented generation (RAG) in chemistry by dynamically retrieving relevant contexts for each question from shared, large-scale corpora. 
However, there is still a lack of scientific context understanding datasets that cover multiple disciplines, modalities, and capabilities. To fill this gap, this work introduces a comprehensive dataset for evaluating the scientific context understanding capability of LLMs.

\section{Datasets} 
This section presents the dataset construction process in \OURS{}, which involves formulating evaluation competencies, collecting scientific data, generating questions and answers, and conducting rigorous verification.

\subsection{Evaluation Competencies} \label{sec:competencies}
Inspired by the ability definition in \cite{chen2024benchmarking}, we formulate four capabilities essential for evaluating LLMs in scientific contexts:
\begin{itemize}[left=0pt]
    \item \textbf{Relevant Information Identification}: LLMs must effectively distinguish between relevant information and extraneous noise within complex scientific contexts. In real-world scenarios, scientific data often contains contextually related but non-essential information. Robust models should be able to filter out such noise to ensure accurate understanding and responses. 
    \item \textbf{Information-absence Detection}: The ability to abstain from responding when all contextual data is irrelevant or unreliable. Scientific queries often require precise evidence,  and when no valid information is present in the context, LLMs should refrain from generating speculative or hallucinated responses.
    \item \textbf{Multi-source Information Integration}: Scientific queries often require synthesizing data from multiple sources. LLMs must aggregate and compare information across different contextual segments to generate precise and contextually grounded answers.
    \item \textbf{Context-aware Inference}: The capability to perform logical inference based on context. Since context in scientific domains may be fragmented or incomplete, LLMs need to analyze relationships, deduce implicit knowledge, and generate accurate answers.
\end{itemize}
These four competencies form the foundation of \OURS{} and provide a systematic evaluation framework for context understanding in scientific domains.

\subsection{Source Data Collection}

\paragraph{Scientific Domains}
To evaluate LLMs in scientific contexts comprehensively, we curate data from diverse scientific domains, including Biology, Chemistry, Physics, Biomedicine, and Materials Science. These disciplines are fundamental to modern science, encompassing a wide range of knowledge from theoretical principles to experimental data, ensuring a broad and representative assessment of long-context understanding capabilities in scientific applications.

\paragraph{Data Modalities}
To support a broad evaluation of scientific context understanding capabilities, we consider three distinct data modalities: (1) {Unstructured Text}, (2) {Structured Tables}, and (3) {Semi-structured Knowledge Graphs} (KGs). Each modality presents unique challenges, enabling a holistic assessment of LLMs' retrieval, synthesis, reasoning, and integration capabilities in scientific domains.

Specifically, unstructured text corpora consist of scientific literature, allowing LLMs to retrieve, synthesize, and infer domain knowledge from textual sources. We collect thousands of recent research papers and experimental protocols from open-access repositories such as arXiv.
Structured tables contain numerical and categorical data, testing LLMs' capacity to interpret structured knowledge, recognize contextual dependencies, and perform quantitative reasoning. We collect nuclear data from IAEA\footnote{\url{https://www-nds.iaea.org}}, material properties from Material Project\footnote{\url{https://next-gen.materialsproject.org}}, and molecular and protein properties from PubChem\footnote{\url{https://pubchem.ncbi.nlm.nih.gov}}.
Knowledge graphs encode scientific knowledge as interconnected entities and relational networks, enabling the assessment of LLMs' abilities in relational inference, hierarchical knowledge traversal, and cross-domain knowledge synthesis. We collect well-established scientific KGs, including Gene Ontology\footnote{\url{https://geneontology.org}} for gene-function relationships, HIPPIE \cite{alanis2016hippie} for protein-protein interactions, PharmKG \cite{zheng2021pharmkg} for drug-target interactions, and PrimeKG \cite{chandak2023building}  for clinical entity relationships. 

\begin{figure}[t]
    \centering
    \vspace{-0.2cm}
    
    \includegraphics[width=0.48\textwidth]{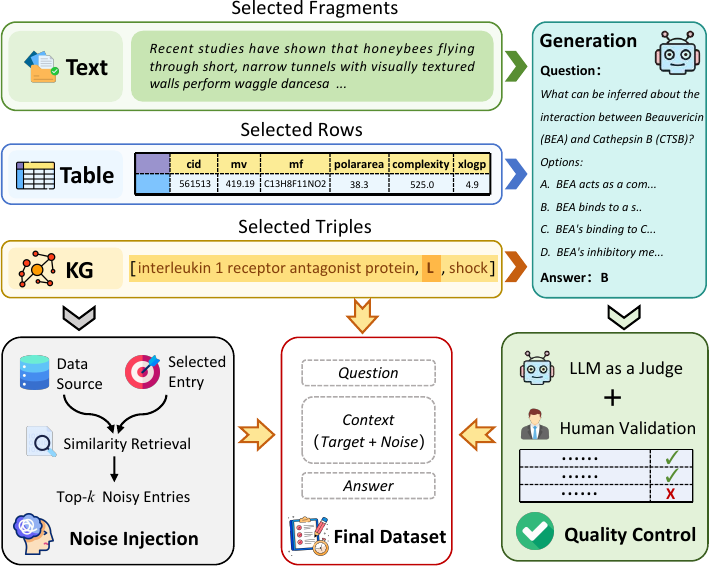} 
    \caption{Illustration of data generation pipeline in \OURS{}, mainly consisting of question generation, noise injection, and quality control.}
    \label{fig:Data_Construction} 
    \vspace{-0.4cm}
\end{figure}

\subsection{Data Generation}
Building on the collected source data, we construct corresponding datasets tailored to assess the proposed four competencies outlined in Sec. \ref{sec:competencies}. The data generation pipeline is illustrated in Figure \ref{fig:Data_Construction}, which involves (1) question generation, (2) noise injection, and (3) quality control.

\paragraph{Question Generation}
We first sample a subset of texts, table rows, or KG triples from the full databases,
\begin{equation}
        \mathcal{D} = \{ d_i \mid d_i = \phi(\mathcal{S}),\ i = 1, 2, \ldots, N \},
\end{equation}
where $\mathcal{S}$ denote the large-scale scientific data source, and $\phi$ is the sampling operation. $d_i$ is a single data record or a small set of related entries. Given each selected data unit $d_i$, the question generation process is defined as:
\begin{equation}
    (q_i, a_i) = f_\text{LLM}(p \oplus d_i),
\end{equation}
where $p$ is a manually designed prompt, and $\oplus$ denotes string concatenation. $f_\text{LLM}$ represents the LLM that generates semantically diverse and contextually relevant questions $q_i$ and corresponding answers $a_i$ based on $d_i$. For each evaluation competency, we craft the prompt to ensure the questions are reasonable and aligned with the required competencies. The detailed prompts are provided in Appendix~\ref{ap:construct_prompt}. 
These generated questions span various formats, including Q\&A, multiple-choice, content completion, and true/false validation, offering a robust assessment of context understanding abilities.

\begin{table*}[t]
    \vspace{-1em}
    \small
    \centering
    \caption{Statistic of the \OURS{} dataset, which comprises ten sub-datasets derived from diverse scientific data. The detailed data sources are listed in Appendix \ref{ap:data_source}.}
    \vspace{-0.5em}
    
    \renewcommand{\arraystretch}{1.3} 
    \resizebox{1\linewidth}{!}{ 
        \begin{tabular}{lllcccccc}
            \toprule
            \textbf{Sub-dataset} & \textbf{Domain} & \textbf{Source} & \textbf{Modality} & 

            \textbf{\#Info. Indent.} & \textbf{\# Abs. Detec.} & \textbf{\# Info. Integ.} & \textbf{\# Con. Infer.} & \textbf{\# Total} \\
            \midrule
            MatText & Materials & arXiv & Text & 216 & 146 & 222 & 356 & 940\\
            BioText & Biology & Biorxiv & Text & 236& 97& 318& 317& 968\\
            MatTab & Materials & Material Project & Table & 299& 150& 287& 200& 936\\
            IaeaTab & Physics & IAEA & Table & 442& 222& 286& 180& 1130\\
            ProtTab & Biology & Pubchem & Table & 496& 249& 327& 180& 1252\\
            MolTab & Chemistry & Pubchem & Table & 516& 259& 350& 180& 1305\\
            GoKG & Biology & Gene Ontology & KG & 507& 254& 239& 180& 1180\\
            HipKG & Biology & HIPPIE & KG & 470& 236& 319& 140& 1165\\
            PhaKG & Biomedicine & PharmKG & KG & 512& 256& 281& 168& 1217\\
            PriKG & Biomedicine & PrimeKG & KG & 410& 205& 382& 253& 1250\\
            \bottomrule
        \end{tabular}
    }
    \label{tab:dataset_overview} 
    \vspace{-.5em}

\end{table*}

\paragraph{Noise Injection}

Following question generation, we extract noisy information from the source data and inject them into the context. Specifically, we inject semantically similar yet unrelated entries into the context using an embedding-based similarity search. Formally, each sample before noise injection is denoted as \( x_i = (q_i, a_i, d_i) \).
To select distractor entries, we first compute embeddings for all candidate entries in the dataset \( \mathcal{D} \) using Sentence-BERT~\cite{reimers2019sentence}.

\begin{equation}
\mathbf{h}_{d_j} = f_\text{S-BERT}(d_j), \quad \forall d_j \in \mathcal{D},
\end{equation}
where $f_\text{S-BERT}$ denotes the embedding function.
We then employ the cosine similarity to efficiently retrieve the Top-$k$ entries most similar to the selected entry $d_i$:
\begin{equation}
\mathcal{N}_i = \underset{d_j \in \mathcal{D} \setminus \{d_i\}}{\operatorname{Top}\text{-}k} \; \text{sim}\left(\mathbf{h}_{d_i}, \mathbf{h}_{d_j}\right),
\end{equation}
where $\mathrm{sim}(\cdot, \cdot)$ denotes the cosine similarity between embedding vectors. 
The final sample after noise injection is represented as \( \tilde{x}_i = (q_i, a_i, d_i \oplus \mathcal{N}_i) \), where \( \mathcal{N}_i \) contains the \( k \) selected distractor entries used to augment the context. We sample \( k \in [200, 300] \) for structured tables and KGs, and set \( k = 5 \) for unstructured text.

Through this approach, the injected noise closely mimics the type of confusing or misleading information that LLMs may encounter in practice, ensuring the benchmark dataset remains both challenging and realistic.

\paragraph{Quality Control}
To maintain the rigor of the constructed dataset, we implement a two-stage verification process to ensure data quality: 
\begin{itemize}[left=0pt]
    \item \textit{LLM as a Judge}. we used advanced LLMs (e.g., GPT-4o) as automated evaluators to check if each answer is directly extractable or logically deducible from the provided context, using a clear prompt. Only instances marked "Yes" were kept.
    
    \item \textit{Human Expert Validation}. Domain experts then manually reviewed the filtered data based on three aspects: (1) whether the question tests the intended competency, (2) whether the question is expressed clearly and logically, and (3) weather the answer is fully supported by contexts and factually correct.  Only instances that received a "Yes" for all three criteria were accepted.
\end{itemize}
As a result, 90.83\% of instances in our dataset met the high-quality criteria. Detailed information about data quality control can be found in Appendix \ref{ap:data_quality}.

\subsection{The Final Dataset}

Based on the data collection, generation, and quality control processes described above, we construct the final \OURS{} dataset, encompassing ten distinct sub-datasets (two unstructured text datasets, four structured table datasets, and four knowledge graph datasets), covering diverse scientific fields.
Each sub-dataset contains approximately a thousand high-quality questions, leading to a total of 11,343 questions across the entire dataset. 
An overview of the dataset composition is presented in Table \ref{tab:dataset_overview}, summarizing the scientific data source, modality, and question distribution for each sub-dataset. Additionally, representative examples are provided in Appendix \ref{ap:QA_examples}.

\begin{table*}[t]
 
\caption{Performance of LLMs across ten sub-datasets on \OURS{}. \underline{Underline results} indicate the best results among all models. \textbf{Bold results} indicate the best results in each category.}
\vspace{-0.5em}
\resizebox{\textwidth}{!}{%

\begin{tabular}{ccccccccccc>{\columncolor{lightgray}}c}
\toprule
\textbf{Models} & \textbf{MatTab} & \textbf{IaeaTab}& \textbf{MolTab} &\textbf{ProtTab} & \textbf{PhaKG} & \textbf{PriKG} & \textbf{HipKG} & \textbf{GoKG} & \textbf{BioText} &\textbf{MatText}  &\textbf{Overall}\\ 
\midrule
GPT-4o& \textbf{{68.79}}& \textbf{{56.55}}& 55.79& \textbf{{52.64}}& \textbf{{55.71}}& \textbf{{54.80}}& 68.50& 74.32&   79.03&64.57&\textbf{{61.52}}\\
GPT-4o-mini& 40.71&38.85&46.67&44.57&40.59&52.64&65.20&73.14&\textbf{\underline{79.24}}&\textbf{\underline{65.00}}&54.57\\
Claude-3.5-Sonnet  & 48.48& 42.03& \textbf{{67.91}}& 52.22& 50.94& 45.96& \textbf{\underline{75.78}}& \textbf{\underline{84.07}}&58.06&61.49&59.20\\

\midrule
DeepSeek-R1& \textbf{\underline{73.71}}  & \textbf{\underline{71.89}} & \textbf{\underline{74.69}}  &  \textbf{\underline{72.44}} & \textbf{\underline{58.66 }} & \textbf{\underline{58.20}}  & 69.66  &  \textbf{{79.18}} & \textbf{74.79}  & \textbf{63.09}  & \textbf{\underline{69.72}}  \\

Qwen3-8B&  63.14  &  59.20 &  70.80 & 69.33   & 55.16 & 54.48  & \textbf{74.68}   & 73.98  &   69.73  &   55.11 & 64.69 \\

DeepSeek-V3&{56.62}& 54.07&                              {59.85}& {52.08}&                             {52.18}& {51.92}& {63.42}& {72.29}&66.74&                                45.31 &{57.50}\\
Llama4-Maverick& 46.47  &   47.79 & 48.20  &  43.61 &  48.32 & 49.28  &  64.81 &   72.71&  63.02 & 54.15  &  53.65 \\

Llama4-Scout&  48.93  &  47.70 & 46.90  &  46.17  & 39.77  &  48.08 & 59.57   & 66.27 &  61.88  &   48.51 & 51.16 \\

 Llama3.1-70B-it& 38.25& 39.73& 44.44& 41.29& 44.70& 44.00& 59.31& 70.17& 66.53& 51.91&49.80\\
 Qwen2.5-7B-it         & 28.10& 32.65& 43.30& 39.46& 36.15& 45.60& 53.99& 62.46&                             68.18&                                 {59.68}&46.62\\

 GLM4-9B-Chat& 31.41& 25.84& 47.82& 43.45& 36.03& 44.56& 57.94& 60.51& 67.77& 50.96&46.46\\
Llama3.1-8B-it&                                    28.85& 34.34&                              42.76& 39.78&                             38.29& 46.56& 52.62& 59.32&                             64.26&                                 49.36&45.50\\
 Gemma2-9B-it& 32.91& 32.21& 42.91& 37.22& 37.39& 50.48& 56.57& 57.29& 37.77& 29.67&42.21\\

 Ministral-8B-it& 23.08& 19.12& 35.56& 37.38& 22.76& 37.92& 48.51& 52.88& 48.14& 45.32&37.58\\

 \midrule

 ChemDFM-v1.5-8B& \textbf{33.65}& \textbf{31.15}& \textbf{35.56}& \textbf{36.82}& \textbf{40.43}& \textbf{30.72}& \textbf{49.70}& \textbf{56.44}& 26.11& 18.91&\textbf{36.80}\\
 SciGLM-6B& 11.86& 11.50& 17.70& 14.94& 19.56& 20.88& 21.46& 28.31& \textbf{44.17}& \textbf{31.35}&21.58\\
 LlaSMol-Mistral-7B& 13.35& 12.83& 16.55& 14.70& 21.54& 19.84& 22.83& 29.92& 33.13& 20.98&20.42\\
  ChemLLM-7B-chat& 3.42& 6.02& 8.81& 8.15& 13.45& 5.92& 5.15& 15.51& 39.94& 22.67&12.16\\

\bottomrule

\end{tabular}%
}
\vspace{-.5em}
\label{tab:dataset_comparison}
\end{table*}

\section{Experiments}
In this section, we evaluate the performance of various LLMs on \OURS{}, and provide a thorough analysis of their capabilities in understanding scientific contexts.

\subsection{Experimental Setup}
\paragraph{Models} We select 18 advanced LLMs, including 3 proprietary models (GPT-4o \cite{openai2024gpt4ocard}, Claude-3.5-Sonnet \cite{anthropic2024claude}, GPT-4o-mini), 11 open-source general-purpose models (DeepSeek-V3 \cite{deepseekai2024deepseekv3technicalreport},  DeepSeek-R1 \cite{guo2025deepseek}, Qwen2.5-7B-Instruct \cite{qwen2025qwen25technicalreport}, Qwen3-8B (with explicit thinking) \cite{yang2025qwen3technicalreport}, Llama3.1-8B-Instruct, Llama3.1-70B-Instruct \cite{dubey2024llama}, Llama-4-Maverick-17B-128E-Instruct, Llama-4-Scout-17B-16E-Instruct \cite{meta_llama4} , Ministral-8B-Instruct \cite{jiang2023mistral}, GLM4-9B-Chat \cite{glm2024chatglmfamilylargelanguage}, Gemma2-9B-it \cite{gemmateam2024gemma2improvingopen}, 4 open-source scientific-domain models (SciGLM-6B \cite{zhang2024sciglm}, LlaSMol-Mistral-7B \cite{yu2024llasmol}, ChemLLM-7B-Chat \cite{zhang2024chemllm}, ChemDFM-v1.5-8B \cite{zhao2024chemdfm}).
For detailed information about these models, please refer to Appendix \ref{ap:models}.

\paragraph{Settings} 
To ensure a fair evaluation across all models, we adopt a unified prompting template that standardizes input formatting. Specifically, each input consists of a system prompt that specifies the question type and defines answer format requirements, contexts, and a question designed to assess one of the four core competencies in \OURS{}.
Given that each question in \OURS{} has a deterministic answer, we adopt \textit{accuracy} as the evaluation metric for all question types across the tasks of relevant information identification, multi-source information integration, and context-aware inference. For the task of information-absence detection, we use the \textit{rejection rate} as the evaluation metric.

\subsection{Overall Results}

Table \ref{tab:dataset_comparison} shows the performance of 18 LLMs on \OURS{} across ten sub-datasets. The results highlight several important trends.
First, models with explicit reasoning capabilities demonstrate clear advantages. The reasoning-augmented open-source model DeepSeek-R1 achieves the highest overall accuracy, outperforming both proprietary models (e.g., GPT-4o ) and other general-purpose open-source models. Qwen3-8B with explicit thinking also performs strongly, ranking second among open-source models. This indicates that incorporating structured reasoning pathways, even without domain-specific pretraining, can significantly enhance performance in scientific tasks. 
Second, proprietary models such as GPT-4o and Claude-3.5-Sonnet maintain competitive performance, especially in unstructured text-based domains (e.g., BioText, MatText), benefiting from their superior language understanding and generalization capabilities. 
Third, scientific-domain LLMs such as ChemDFM-v1.5-8B and SciGLM-6B exhibit substantially lower performance across all datasets. Although designed for scientific domains, these models tend to lack general reasoning capacity and robustness across modalities. 
Fourth, there is a strong positive correlation between model size and effectiveness. Large-scale models (e.g., GPT-4o,  Llama4-Maverick, and Llama3.1-70B) consistently outperform their smaller counterparts (e.g., GPT-4o-mini, Llama4-Scout, and Llama3.1-8B) across most domains. 

\begin{figure*}[htbp!] 
    \vspace{-1em}

    \includegraphics[width=0.98\textwidth]{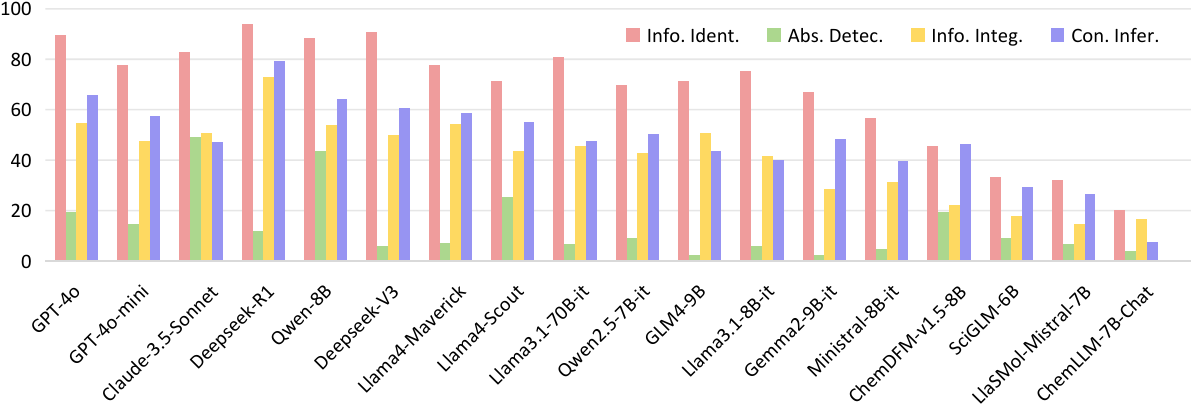}
    \vspace{-0.7em}
    \caption{Performance of LLMs across four competencies on \OURS{}. }    \label{fig:Competencies_comparison}
    \vspace{-.5em}
\end{figure*}

\subsection{Evaluation Results of Four Competencies}
\paragraph{Relevant Information Identification} 
This competency measures a model's ability to locate and select the correct pieces of information from the provided context. As shown in Figure~\ref{tab:Competencies_comparison}, DeepSeek-R1 leads all of the evaluated models, suggesting that explicit reasoning mechanisms effectively enhance factual grounding. DeepSeek-V3, GPT-4o, and Qwen-8B also exhibit strong performance, showing the advantages of in-context retrieval capabilities. In contrast, scientific-domain LLMs exhibit notably weaker performance in identifying relevant contexts across diverse scenarios.

\paragraph{Information-absence Detection}
This metric evaluates whether a model appropriately withholds an answer when the required information is absent. Claude-3.5-Sonnet and Qwen-8B demonstrate relatively high accuracy, suggesting their conservative answering strategy and stronger understanding of uncertainty.  Most models struggle in this competency, with scores below 20\%, indicating a tendency to hallucinate answers when uncertain. This highlights the risk of "overconfidence" in current models, which may pose potential safety risks in the scientific domain.

\paragraph{Multi-source Information Integration}
This task assesses a model's ability to synthesize information from multiple entries to construct a complete answer. DeepSeek-R1 achieves the highest performance, followed by GPT-4o and Llama4-Maverick, suggesting that these models are better equipped to combine multiple data points into coherent and accurate answers.
Among smaller open-source models, GLM4-9B shows a competitive score, even surpassing DeepSeek-V3 in this competency. However, scientific LLMs significantly lag behind, indicating that while these domain-specific models are adept at handling scientific text, they face challenges in effectively synthesizing information from diverse sources.
   
\paragraph{Context-aware Inference} 
This capability reflects a model’s ability to reason over contextually relevant information. DeepSeek-R1 achieves the highest performance, and GPT-4o and Qwen3-8B also perform well, indicating that large-scale models and those enhanced with explicit thinking benefit significantly in contextual reasoning tasks. In contrast, models like Claude-3.5-Sonnet and DeepSeek-V3 show moderate capabilities but fall behind on deeper inference. Scientific-domain models such as ChemLLM-7B-Chat and SciGLM perform poorly, indicating limited general reasoning capabilities despite domain specialization.

\begin{figure*}[htbp!] 
    \includegraphics[width=0.98\textwidth]{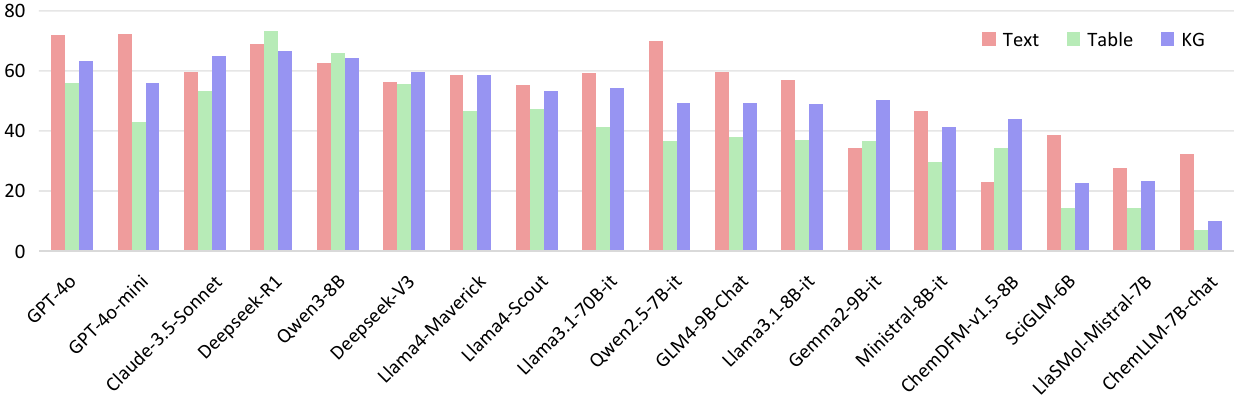}
    \vspace{-0.7em}
    \caption{Performance of LLMs across three modalities on \OURS{}. }    \label{fig:modality_comparison}
    \vspace{-.5em}
\end{figure*}

\subsection{Evaluation Results of Three Modalities}
Figure~\ref{fig:modality_comparison} shows the performance of LLMs across three modalities: \textbf{Text}, \textbf{Table}, and \textbf{KG}, highlighting their strengths and weaknesses in handling diverse scientific data formats.
Overall, LLMs tend to perform best on the text modality, reflecting their strong natural language understanding and generation capabilities. Notably, some smaller models even exceed their average overall performance on text, indicating a relative maturity in handling unstructured text data.
In the table modality, reasoning-augmented models demonstrate a clear advantage, suggesting that explicit reasoning mechanisms and the ability to process structured data significantly benefit table understanding. In contrast, general LLMs show weaker performance on tables, implying challenges in leveraging tabular structure with traditional language modeling approaches.
Similarly, for KG data, models with reasoning enhancements again lead, reflecting their ability to leverage relational and graph-structured information effectively.
Additionally, domain-specific scientific models consistently underperform across all three modalities compared to general-purpose or reasoning-augmented models.

\subsection{Further Discussions}

Our experimental results highlight three key discrepancies in the performance of LLMs on scientific context understanding tasks, underscoring fundamental challenges that require further advancements.

\paragraph{Competency Discrepancy}
The evaluation results reveal notable disparities across the four core competencies. While top-performing models exhibit relatively strong capabilities in identifying relevant information, they struggle with information-absence detection—the ability to abstain from answering when faced with unreliable or insufficient evidence. This suggests that models prioritize generating responses over ensuring accuracy, increasing the risk of hallucinations in scientific applications where factual correctness is critical.
To address this, models should incorporate uncertainty quantification techniques, such as confidence-based rejection mechanisms and calibrated probability outputs, to enhance their ability to detect and reject misleading retrievals. Furthermore, reinforcement learning with human feedback and verification-based prompting strategies could help improve the model’s reliability in rejecting incorrect information.

\paragraph{Modality Discrepancy}
LLMs exhibit relatively better performance on unstructured text compared to structured tables and KGs. This suggests that existing models rely heavily on linguistic patterns and semantic context rather than structured inference and multi-modal data integration. The weaker performance on tables and KGs indicates a bottleneck in structured data comprehension, where models struggle to extract, synthesize, and infer information effectively from unstructured data.
To bridge this gap, models need improved cross-modal alignment, integrating structured data reasoning into their training paradigm. Techniques such as joint pretraining on text, tables, and graphs could enhance structured data understanding.

\paragraph{Specialized vs. General Model Discrepancy}
Although scientific LLMs are explicitly designed for knowledge-intensive tasks, our evaluation shows that they often fail to outperform general-purpose models on our dataset. This suggests that current specialized models lack sufficient reasoning depth, robustness, and flexibility to fully leverage domain knowledge in complex scientific contexts. Their narrower training scope may limit generalization across diverse data modalities and reasoning challenges.
To improve their contextual understanding, scientific models should incorporate targeted fine-tuning using curated scientific evidence and adopt domain-aware prompt engineering strategies. These approaches can help balance deep specialization with the adaptability required to tackle a broad range of scientific tasks, enhancing their effectiveness across diverse scenarios.

\section{Conclusion}
In this work, we introduced \OURS{}, a comprehensive dataset for evaluating context understanding capabilities in large language models within scientific domains. \OURS{} encompasses multiple data modalities (structured tables, knowledge graphs, and unstructured text), spanning diverse scientific disciplines. By systematically assessing four key competencies (Relevant Information Identification, Information-absence Detection, Multi-source Information Integration, and Context-aware Inference), we provide a unified framework to quantify how effectively LLMs perform on science-intensive tasks.
Our experimental findings reveal that, despite notable progress, existing models encounter substantial challenges in accurately interpreting scientific data.  The primary challenge lies in the inherent complexity of scientific data, particularly structured formats such as tables and knowledge graphs, which demand high specialization, precise contextual understanding, and the ability to synthesize fragmented and implicitly related information. 
Even state-of-the-art LLMs show limitations in fully mastering these skills, underscoring the need for significant advancements to enhance their scientific context understanding.

\section*{Limitations}
While \OURS{} offers a comprehensive dataset for evaluating LLMs in scientific contexts, it has several inherent limitations. First, the dataset is predominantly text-based, omitting important non-textual modalities such as images and 3D molecular structures, which play a vital role in many scientific domains. Integrating more multimodal data would enable a more complete and nuanced assessment of models’ context understanding capabilities.
Second, although \OURS{} spans multiple scientific disciplines, it cannot fully capture the extensive heterogeneity of scientific knowledge. As a result, model performance on \OURS{} may not entirely generalize to rapidly evolving scientific fields.
The next version of \OURS{} will aim to include additional datasets from a wider spectrum of scientific areas to enhance coverage and applicability.

\bibliography{ref}

\clearpage
\appendix

\section*{Appendix}

\section{More Results on \OURS{}}\label{ap:additional_results}

Table \ref{tab:Competencies_comparison} and \ref{tab:modality_comparison} present the quantitative evaluation results of LLMs across four competencies and three modalities on \OURS{}, respectively. Table \ref{tab:detail_results} presents the more detailed results of \OURS{}. Table \ref{tab:without-RAG} shows the performance comparison between direct answering and answering with context. The results demonstrate that the integration of context consistently enhances performance.

\begin{table}[htbp]

    \centering
    \caption{Evaluation results of LLMs across four competencies on \OURS{}. 
    \underline{Underline results} indicate the best results among all models. \textbf{Bold results} indicate the best results in each category.
    }
    \renewcommand{\arraystretch}{1} 
    
    \resizebox{\linewidth}{!}{ 
         \begin{tabular}{cccccc}
            \toprule
            \multirow{2}{*}{\textbf{Models}}& \textbf{Info.} & \textbf{Abs.} & \textbf{Info.} & \textbf{Con.} & \multirow{2}{*}{\textbf{Overall}} \\
 & \textbf{Ident.}& \textbf{Detec.}& \textbf{Integ.}& \textbf{Infer.}&\\
            \midrule
            GPT-4o&\textbf{89.72} &19.51 & \textbf{{54.90}} & \textbf{{65.97}}&\textbf{{61.52}}
\\
            GPT-4o-mini& 77.81 &14.71 & 47.54& 57.68&54.57
\\
            Claude-3.5-Sonnet&82.95 & \textbf{\underline{49.10}} & 50.85&47.29& 59.20\\

            \midrule
            DeepSeek-R1&94.13&11.75&  \textbf{\underline{72.78 } }&\textbf{\underline{79.44}}&\textbf{\underline{69.72}}\\
            Qwen3-8B& 88.29 & \textbf{43.57} & 53.87 & 64.38& 64.69 \\

            DeepSeek-V3& \textbf{\underline{90.80}} & 6.05 & 49.80 &{60.53}&{57.50}
\\
Llama4-Maverick&  77.61  &  7.16  &  54.16  &   58.52 &  53.65   \\
Llama4-Scout& 71.38   &  25.42 &   43.65  &  54.95 & 51.16   \\
            Llama3.1-70B-it&  81.05& 6.87&45.44 &47.76&49.80
\\
            Qwen2.5-7B-it&69.92  &{9.02} & 42.95& 50.34& 46.62
\\
            GLM4-9B& 71.48& 2.53& {50.78} &43.82&46.46 \\
            Llama3.1-8B-it&  75.34&5.88 &41.47 &39.99 &45.50
\\
            Gemma2-9B-it &66.97 &  2.38&28.74 &48.22&42.20
\\
            Ministral-8B-it&  56.80&4.76 & 31.32 &  39.62&37.58\\

            \midrule
            ChemDFM-v1.5-8B& \textbf{45.49} &  \textbf{19.31} & \textbf{22.23} & \textbf{46.40} &\textbf{36.80}
\\
            SciGLM-6B&33.24 & 9.01 &18.00 & 29.48&21.58
\\
            LlaSMol-Mistral-7B& 31.96& 6.83 & 14.63 &26.59&20.42
\\
            ChemLLM-7B-Chat& 20.29 & 4.09 & 16.85&7.57&12.16\\

            \bottomrule
        \end{tabular}
    }

    \label{tab:Competencies_comparison}
\end{table}

\begin{table}[htbp]
\small
\caption{Evaluation results of LLMs across three modalities on \OURS{}. \underline{Underline results} indicate the best results among all models. \textbf{Bold results} indicate the best results in each category.}

\resizebox{0.48\textwidth}{!}{%
\renewcommand{\arraystretch}{1.1} 

\begin{tabular}{ccccc}
\toprule
\textbf{Models} & \textbf{Text}& \textbf{Table}& \textbf{KG}&\textbf{Overall}\\ 
\midrule
GPT-4o& 71.91& \textbf{55.91}& 63.13 &\textbf{{61.52}}\\
GPT-4o-mini& \textbf{\underline{72.22}}& 42.98& 55.84&54.57\\
Claude-3.5-Sonnet  & 59.75& 53.41& \textbf{64.99} &59.20\\

\midrule
DeepSeek-R1&69.00&\textbf{\underline{73.19}}&\textbf{\underline{66.64}}&\textbf{\underline{69.72}}\\
Qwen3-8B& 62.53 & 66.02 &  64.27 & 64.69 \\

DeepSeek-V3&56.18& 55.68&{59.77}&{57.50}\\
Llama4-Maverick&  58.65  &  46.51  & 58.54   &53.65\\
Llama4-Scout&55.29&  47.31  & 53.22   &51.16\\

 Llama3.1-70B-it& 59.33& 41.19& 54.30&49.80\\

Qwen2.5-7B-it&\textbf{69.93}& 36.58& 49.38&46.24\\
 GLM4-9B-Chat&59.49&37.94&49.46&46.46\\
Llama3.1-8B-it&56.92& 37.08&49.06&45.50\\
 Gemma2-9B-it&34.27&36.73&50.23&42.21\\
  Ministral-8B-it& 46.75& 29.50& 41.24&37.58\\

 \midrule

 ChemDFM-v1.5-8B& 22.92& \textbf{34.44}& \textbf{44.05}&\textbf{36.80}\\
 SciGLM-6B& \textbf{38.48}& 14.25& 22.51&21.58\\
 LlaSMol-Mistral-7B& 27.74& 14.49& 23.45&20.42\\
  ChemLLM-7B-chat& 32.28& 6.86& 10.01&12.16\\
\bottomrule

\end{tabular}%
}
\label{tab:modality_comparison}
\end{table}

\begin{table*}
    \small
    
 \caption{Detailed evaluation results of LLMs across four competencies on the ten sub-datasets of \OURS}
    \renewcommand{\arraystretch}{1.1}
    \centering
    \resizebox{\linewidth}{!}{ 
    \begin{tabular}{c c c c c c | c c c c c}
        \toprule
        \multirow{2}{*}{Models} & \multicolumn{5}{c|}{MatTab} &\multicolumn{5}{c}{IaeaTab}\\
        \cline{2-11}
         & Info. Ident. & Abs. Detec. & Info. Integ. & Con. Infer. & All & Info. Ident. & Abs. Detec. & Info. Integ. & Con. Infer.& All\\

        \midrule
        GPT-4o & 88.16 & 58.00 & 29.97 & 64.00  & 68.79&81.00 & 26.13 & 44.76 & 52.78&56.55\\
        GPT-4o-mini &71.91 & 39.33 & 10.10 & 39.00 & 40.71&57.01 & 5.86 & 37.06 & 37.78&38.85\\
        Claude-3.5-Sonnet &77.21 & 56.46 & 19.57 & 41.00&  48.48 &56.31 & 23.29 & 37.68 & 37.78& 42.03\\
        \midrule
        DeepSeek-R1 & 97.30 & 16.22 & 69.01 & 88.00&73.71 & 91.82 & 16.36 & 76.06 & 84.44 &71.89 \\
        Qwen3-8B & 82.94 & 90.00 & 19.16 & 76.50  & 63.14  & 68.78 & 56.76 & 41.96 & 66.11    &   59.20 \\

        DeepSeek-V3 &93.31 & 28.00 & 24.74 & 69.00&56.62&79.86 & 5.41 & 44.76 & 65.56&54.07 \\

      Llama4-Maverick &69.90 & 9.33 & 23.00 & 73.00  & 46.47&66.29 & 12.61 & 39.51 & 58.89& 47.79\\

        Llama4-Scout&53.85 & 73.33 & 15.68 & 71.00 & 48.93  & 59.05 & 34.23 & 36.36 & 54.44&  47.70\\
        Llama3.1-70B &69.57 & 26.00 & 6.97 & 45.50& 38.25&62.44 & 6.31 & 30.77 & 39.44 &39.73 \\

        Qwen2.5-7B-instruct & 48.83 & 9.33 & 3.14 & 47.00& 28.10  &45.70 & 8.11 & 32.87 & 30.56&32.65  \\
        GLM4-9B-Chat & 52.84 & 2.67 & 18.82 & 39.00 &31.41 &29.64 & 0.45 & 40.56 & 24.44&25.84 \\
        Llama3.1-8B &58.53 & 4.67 & 4.88 & 37.00 & 28.85 &53.85 & 6.31 & 25.52 & 34.44 &34.34 \\
        Gemma2-9B-it &65.55 & 2.00 & 6.97 & 44.50 & 32.91& 50.23 & 2.70 & 16.78 & 48.89 & 32.21\\
        Ministral-8B-it &44.15 & 2.00 & 3.14 & 36.00 & 23.08 &18.78 & 9.01 & 9.79 & 47.22 & 19.12\\

        \midrule
        ChemDFM-v1.5-8B & 42.81 & 10.67 & 9.41 & 72.00 & 33.64 &36.65 & 14.41 & 21.33 & 53.89&31.15 \\
        SciGLM-6B &5.69 & 1.33 & 2.79 & 42.00 &11.86 &11.99 & 0.90 & 3.85 & 35.56&11.50 \\

        LlaSMol-Mistral-7B & 7.69 & 5.33 & 2.79 & 43.00 & 13.35 &16.29 & 6.31 & 1.75 & 30.00 & 12.83\\        
        ChemLLM-7B-Chat &1.67 & 2.67 & 0.00 & 11.50 &3.42 &9.28 & 0.45 & 4.90 & 6.67 &6.02 \\

        \midrule
        \midrule
         \multirow{2}{*}{Models} & \multicolumn{5}{c|}{MolTab} &\multicolumn{5}{c}{ProtTab}\\
        \cline{2-11}
         & Info. Ident. & Abs. Detec. & Info. Integ. & Con. Infer. & All & Info. Ident. & Abs. Detec. & Info. Integ. & Con. Infer.& All\\
        \midrule
        GPT-4o & 91.09 & 9.27 & 30.29 & 71.11 &55.79 & 90.52 & 14.46 & 18.96 & 62.22  & 52.64\\        
        GPT-4o-mini &68.99 & 13.13 & 32.29 & 58.89& 46.67 &72.58 & 9.24 & 19.27 & 62.22  & 44.57\\
        Claude-3.5-Sonnet &92.84 & 58.59 & 42.53 & 60.56 &67.91  & 77.08 & 31.43 & 20.12 & 70.00& 52.22\\
         \midrule
        DeepSeek-R1 & 96.90 & 14.29 & 75.86 & 93.33 &74.69 & 96.77 & 17.74 & 65.43 & 93.33 & 72.44\\

        Qwen3-8B & 85.85 & 62.16 & 46.00 & 88.33& 70.80  & 89.72 & 62.25 & 32.72 & 89.44 &   69.33 \\
         
         DeepSeek-V3 & 94.38 & 5.79 & 39.71 & 77.78&59.85 &96.57 & 6.02 & 22.94 & 46.11& 52.08\\      
         Llama4-Maverick &73.06 & 2.70 & 39.43 & 59.44 & 48.20&71.77 & 1.20 & 22.63 & 62.78 &43.61 \\
        Llama4-Scout &60.47 & 9.27 & 36.57 & 82.22& 46.90 &64.92 & 16.47 & 26.61 & 71.11& 46.17 \\

         Llama3.1-70B &71.90 & 1.93 & 30.29 & 54.44 & 44.44&79.44 & 2.81 & 14.07 & 38.89 &41.29 \\
         Qwen2.5-7B-instruct & 57.75 & 5.41 & 37.71 &68.33& 43.30  &61.69 & 9.64 & 19.88 & 55.00 & 39.46 \\        
         GLM4-9B &66.86 & 0.39 & 50.29 & 56.67&47.82 &60.89 & 2.01 & 44.34 & 51.11& 43.45\\

        Llama3.1-8B-instruct &70.74 & 0.00 & 31.14 & 46.67& 42.76 &67.74 & 2.41 & 21.10 & 48.33 & 39.78\\        
        Gemma2-9B-it&63.57 & 0.39 & 36.86 & 56.67 & 42.91 &62.70 & 0.80 & 18.65 & 51.11 & 37.22 \\
        Ministral-8B-it &51.55 & 1.16 & 31.71 & 46.67 & 35.56 & 59.07 & 1.20 & 23.24 & 53.33& 37.38\\

        \midrule
        ChemDFM-v1.5-8B &38.95 & 33.20 & 22.29 & 55.00& 35.56 & 38.91 & 34.94 & 19.27 & 65.56& 36.82\\
        SciGLM-6B &23.26 & 2.32 & 8.86 & 41.11 & 17.70&19.76 & 3.21 & 6.42 & 33.33 &14.94 \\
        LlaSMol-Mistral-7B &22.48 & 5.79 & 10.00 & 27.78 & 16.55 & 22.18 & 5.62 & 3.98 & 26.11 &14.70 \\ 
        ChemLLM-7B-Chat & 11.63 & 0.00 & 14.00 & 3.33& 8.81 &13.31 & 0.00 & 9.48 & 2.78 & 8.15\\

        \midrule
        \midrule
         \multirow{2}{*}{Models} & \multicolumn{5}{c|}{PriKG} &\multicolumn{5}{c}{HipKG}\\
        \cline{2-11}
         & Info. Ident. & Abs. Detec. & Info. Integ. & Con. Infer. & All & Info. Ident. & Abs. Detec. & Info. Integ. & Con. Infer.& All\\
         \midrule
        GPT-4o & 70.98 & 24.88 & 41.62 & 72.73  &54.80 & 97.02 & 33.90 & 43.26 & 88.57 & 68.50\\
        GPT-4o-mini & 74.88 & 27.80 & 31.15 & 69.17 &52.64 &83.51 & 20.34 & 43.26 & 67.86& 65.20\\
        Claude-3.5-Sonnet &63.66 & 28.29 & 29.66 & 56.13&45.96 & 97.85 & 75.32 & 43.26 & 77.14& 75.78\\
         \midrule
        DeepSeek-R1 &77.45 & 15.69 & 43.16 & 84.13& 58.20&97.44 & 3.39 & 68.35 & 94.29  &69.66 \\

        Qwen3-8B & 79.27 & 30.24 & 35.86 & 62.06 & 54.48  & 94.47 & 42.80 & 57.99 & 98.57 &   74.68 \\
         
         DeepSeek-V3 &73.90 & 1.95 & 40.58 & 73.91 & 51.92&94.47 & 9.32 & 52.66 & 75.00&63.42 \\        

         Llama4-Maverick & 69.76 & 4.39 & 39.79 & 66.80 & 49.28&87.02 & 13.56 & 69.59 & 65.71&64.81 \\
        Llama4-Scout & 77.32 & 20.98 & 28.80 & 51.78  &48.08  &80.43 & 42.80 & 47.65 & 45.00& 59.57 \\

         Llama3.1-70B-it &74.88 & 9.27 & 20.68 & 57.31  & 44.00&91.06 & 14.41 & 42.95 & 65.71& 59.31\\        

         Qwen2.5-7B-it & 68.05 & 22.93 & 27.75 & 54.55 & 45.60  &60.64 &3.39 &48.90 & 62.14& 53.99 \\         
         GLM4-9B-Chat &70.98 & 6.83 & 27.23 & 58.50 &44.56 &85.32 & 5.51 & 52.98 & 65.71 & 57.94\\

        Llama3.1-8B-it &75.37 & 18.54 & 27.75 & 50.99  & 46.56 & 82.55 & 10.59 & 45.14 & 40.00& 52.62\\
        Gemma2-9B-it&78.78 & 6.83 & 30.37 & 70.36 & 50.48 &91.49 & 3.81 & 36.99 & 72.86 & 56.57\\
        Ministral-8B-it &55.12 & 13.66 & 23.04 & 52.17 &37.92 &66.60 & 1.27 & 31.03 & 47.50 & 48.51\\

        \midrule
        ChemDFM-v1.5-8B & 42.20 & 34.15 & 10.47 & 39.92&30.72 &50.85 & 72.46 & 14.73 & 87.14&49.70 \\        
        SciGLM-6B &42.93 & 4.39 & 4.71 & 22.92& 20.88&33.19 & 1.27 & 5.33 & 52.86&21.46\\
        LlaSMol-Mistral-7B &33.90 & 10.73 & 5.24 & 26.48& 19.84& 26.38 & 8.47 & 9.72 & 65.00 & 22.83\\ 

        ChemLLM-7B-Chat &8.05 & 3.41 & 7.59 & 1.98 & 5.92& 3.62 & 1.27 & 8.15 & 10.00 & 5.15\\

        \midrule
        \midrule
      
    \end{tabular}
    }

    \label{tab:detail_results}
\end{table*}

\begin{table*}
    \small
    \renewcommand{\arraystretch}{1.1}
    \centering
    \resizebox{\linewidth}{!}{ 
    \begin{tabular}{c c c c c c | c c c c c}
        \toprule
        \multirow{2}{*}{Models} & \multicolumn{5}{c|}{GoKG} &\multicolumn{5}{c}{PhaKG}\\
        \cline{2-11}
         & Info. Ident. & Abs. Detec. & Info. Integ. & Con. Infer. & All & Info. Ident. & Abs. Detec. & Info. Integ. & Con. Infer.& All\\
        \midrule
        GPT-4o & 91.91 & 11.02 & 87.45 & 96.67 &74.32 &88.09 & 16.80 & 45.91 & 32.74 & 55.71\\
        GPT-4o-mini &90.34 & 23.23 & 76.57 & 90.56&73.14 &63.87 & 8.20 & 38.08 & 23.21 & 40.5\\
        Claude-3.5-Sonnet &91.91 & 82.28 & 70.71 & 82.22 &84.07 &88.77 & 28.63 & 64.00 & 14.88 &50.94 \\
        \midrule
        DeepSeek-R1 &93.65 & 23.81 & 94.92 & 95.56& 79.18& 93.33 & 10.00 & 51.47 & 47.62& 58.66\\

        Qwen3-8B & 94.87 & 29.92 & 74.48 & 76.67  & 73.98  & 89.26 & 26.95 & 37.01 & 29.50   &  55.16  \\
        
        DeepSeek-V3 &91.72 & 1.57 & 88.70 & 95.56&72.29 &86.91 & 0.39 & 46.62 & 34.52 & 52.18\\

       Llama4-Maverick & 90.93 & 7.09 & 90.38 & 90.56 & 72.71&73.44 & 9.38 & 47.33 & 32.74& 48.32\\
        Llama4-Scout & 85.60 & 36.22 & 37.66 & 92.22 & 66.27 &56.84 & 16.41 & 40.57 & 22.02 &  39.77 \\
        Llama3.1-70B-it &92.70 & 3.54 & 74.48 & 95.00 & 70.17&72.07 & 2.73 & 41.28 & 30.95 & 44.70\\

         Qwen2.5-7B-it & 88.17 & 12.60 & 45.61 & 82.78& 62.46  &54.30 & 17.19 & 31.32 & 17.86 & 36.15\\
        GLM4-9B-Chat &87.57 & 5.12 & 51.46 & 74.44  & 60.51&62.89 & 2.34 & 32.38 & 15.50 & 36.05\\

        Llama3.1-8B-it &91.32 & 2.76 & 39.75 & 75.00 & 59.32&56.84 & 12.89 & 36.65 & 23.21 & 38.29\\
        Gemma2-9B-it&84.42 & 0.39 & 47.28 & 74.44 & 57.29 &61.13 & 3.12 & 38.43 & 19.00 &  37.39\\
        Ministral-8B-it &82.05 & 4.33 & 27.62 & 72.78 &52.88 &40.43 & 5.47 & 15.66 & 7.14 &22.76 \\

        \midrule

        ChemDFM-v1.5-8B &72.58 & 43.31 & 39.75 & 51.67 &56.44 & 58.01 & 41.80 & 25.27 & 15.00 & 40.43\\
        SciGLM-6B &43.39 & 9.84 & 15.48 & 28.89 & 28.31&40.62 & 3.52 & 3.91 & 5.95 & 19.56\\
        LlaSMol-Mistral-7B &50.69 & 9.45 & 13.39 & 22.22&29.92 & 47.85 & 7.03 & 1.07 & 1.50& 21.54\\ 

        ChemLLM-7B-Chat & 19.53 & 14.96 & 10.46 & 11.67  & 15.51&30.08 & 3.52 & 0.36 & 2.00 & 13.45\\
   
        \midrule
        \midrule

         \multirow{2}{*}{Models} & \multicolumn{5}{c|}{MatText} &\multicolumn{5}{c}{BioText}\\
        \cline{2-11}
         & Info. Ident. & Abs. Detec. & Info. Integ. & Con. Infer. & All & Info. Ident. & Abs. Detec. & Info. Integ. & Con. Infer.& All\\
         \midrule
        GPT-4o & 99.07 & 0.68 & 97.30 & 49.44  & 64.57& 99.58 & 0.00 & 97.48 & 69.40  &79.03 \\
        GPT-4o-mini & 96.76 & 0.00 & 91.44 & 55.90 &65.00 &98.31 & 0.00 & 96.23 & 72.24 &79.24 \\
        Claude-3.5-Sonnet &99.07 & 54.11 & 96.40 & 19.94& 61.49 &84.75 & 52.58 & 84.59 & 13.25 & 58.06 \\
         \midrule
        DeepSeek-R1 &98.61& 0.00 & 92.59 & 48.31&63.09 &96.61 & 0.00 & 90.91 & 65.38&74.79 \\

        Qwen3-8B &  98.61 & 17.12 & 97.30 & 17.98 & 55.11  & 99.15 & 17.53 & 96.23 & 37.22 &  69.73  \\
         
         DeepSeek-V3 &98.15 & 2.05 & 42.34 & 32.87&45.31 &98.73 & 0.00 & 94.97 & 35.02& 66.74\\

           Llama4-Maverick &87.04 & 8.22 & 84.68 & 33.99&54.15 &86.86 & 3.09 & 85.22 & 41.32& 63.02\\
        Llama4-Scout &89.35 & 1.37 & 77.93 & 24.72&  48.51  &86.02 & 3.09 & 88.68 & 35.02& 61.88 \\
        Llama3.1-70B-it &98.15 & 0.68 & 97.30 & 16.57 & 51.59&98.31 & 1.03 & 95.60 & 33.75& 66.53\\

         Qwen2.5-7B-it & 68.05 & 22.93 & 27.75 & 54.55 & 59.68  &60.64 &3.39 &48.90 & 62.14& 66.18 \\
        GLM4-9B-Chat &98.61 & 0.00 & 93.24 & 16.57  &50.96 &99.15 & 0.00 & 96.54 & 36.28& 67.77\\

        Llama3.1-8B-it &98.15 & 0.68 & 89.64 & 14.61&49.36 &98.31 & 0.00 & 93.08 & 29.65 & 64.26\\
        Gemma2-9B-Chat&56.94 & 0.68 & 10.53& 26.12& 29.67 &54.85 & 3.09 & 55.03 & 18.30 & 37.77\\
        Ministral-8B-it &83.33 & 19.18 & 76.58 & 13.48 & 45.32&66.95 & 18.56 & 71.38 & 19.87&48.14 \\

        \midrule
        ChemDFM-v1.5-8B &34.72 & 6.12 & 24.53 & 13.76  &18.91 &39.24 & 16.49 & 35.22 & 10.09& 26.11\\
        ChemLLM-7B-Chat &45.83 & 11.56 & 54.72 & 8.43& 31.35&59.92 & 3.09 & 58.81 & 17.35 & 44.17\\
        LlaSMol-Mistral-7B & 44.91 & 5.44 & 49.06 & 8.71 &20.98 &47.26 & 4.12 & 49.37 & 15.14& 33.13\\ 
        SciGLM-6B &54.63 & 29.25 & 66.04 & 12.92&22.67 &56.96 & 34.02 & 62.58 & 19.24 & 39.94\\
        \midrule
        \midrule

    \end{tabular}

    }

\end{table*}

\begin{table*}
\centering
\small
\caption{Performance comparison on \OURS{}: Direct Answering vs. Answering with Contexts.}

\resizebox{\textwidth}{!}{%
\begin{tabular}{c|cc|cc|cc|cc|cc}
\hline
\multicolumn{1}{c|}{\multirow{2}{*}{Model}} & \multicolumn{2}{c|}{\textbf{MatTab}}          & \multicolumn{2}{c|}{\textbf{IaeaTab}}               & \multicolumn{2}{c|}{\textbf{MolTab}}          & \multicolumn{2}{c|}{\textbf{ProtTab}}           & \multicolumn{2}{c}{\textbf{PhaKG}}                 \\ \cline{2-11} 
\multicolumn{1}{c|}{}                       & \multicolumn{1}{c}{Direct} & \multicolumn{1}{c|}{Context} & \multicolumn{1}{c}{Direct} & \multicolumn{1}{c|}{Context} & \multicolumn{1}{c}{Direct} & \multicolumn{1}{c|}{Context} & \multicolumn{1}{c}{Direct} & \multicolumn{1}{c|}{Context} & \multicolumn{1}{c}{Direct} & \multicolumn{1}{c}{Context} \\ \hline
GPT-4o& 14.64& {{68.79}}& 15.31& {{56.55}}& 26.82& 55.79& 23.64& {{52.64}}& 16.81& {{55.71}}\\
GPT-4o-mini&{15.38}&40.71&18.67&38.85&25.52&46.67&24.84&44.57&14.01&40.59\\
Claude-3.5-Sonnet&15.22&48.48&{{23.45}}&42.03&{{32.95}}&{{67.91}}&{{31.07}}&52.22&{26.62}&50.94\\

\midrule
DeepSeek-R1 & 6.34  &  {{73.71}} & 16.01  & {{71.89}}  &  12.61   & {{74.69}}  & 10.26   & {{72.44}} &  11.51  & {{58.66}}  \\
Qwen3-8B &14.10  & 63.14   & 9.12 & 59.20   & 13.95   & 70.80  & 13.26   &   69.33 &   11.85 &  55.16 \\
DeepSeek-V3&14.82& 56.62 &  22.65  & 54.07 & 31.88  & 59.85 &  {26.20}  & {52.08} &  15.80  &  {52.18} \\

Llama4-Maverick & {{25.53}}  & 46.47  & 20.35  &   47.79  &  35.33   & 48.20   & 39.70   & 43.61  &   18.32 & 48.32  \\
Llama4-Scout &25.43 & 48.93   & {{28.94}} & 47.70   & {{46.21}}   & 46.90   & {{39.78}}   & 46.17  &   {18.41}  &  39.77 \\

Llama3.1-70B-it&10.04&38.25&16.19  &39.73& 21.30  & 44.44 &  22.60  & 41.29 & 13.15   & 44.70  \\
Qwen2.5-7B-it&14.64&28.10&18.94&32.65&21.07&43.30&20.69&39.46&15.93&36.15\\
GLM4-9B-Chat &12.61& 31.41 &16.73& 25.84 &22.53& 47.82 &22.28& 43.45 &13.53&  36.03 \\

Llama3.1-8B-it& 14.21&28.85&14.78&34.34&18.54&42.76&17.97&39.78&{16.35}&38.29\\
Ministral-8B-it&0.82& 23.08 &8.29& 19.12 &8.00& 35.56 &4.66& 37.38 &15.05& 22.76  \\
Gemma2-9B-it &13.25    & 32.91 &  10.27  & 32.21 &   12.87 & 42.91 &   13.74 & 37.22 &  11.45  &  37.39 \\


\midrule
ChemDFM-v1.5-8B&14.38& {33.65} &{13.54}& {31.15} &{26.36}& {35.55} &{28.63}& {36.82} &{{44.53}}&  {40.43} \\
SciGLM-6B&12.18& 11.86 &10.44& 11.50 &15.56& 17.70 &13.58& 14.94 &13.56&19.56   \\
LlaSMol-Mistral-7B&11.97& 13.35 &10.88& 12.83 &13.71& 16.55 &11.98&14.70  &23.62& 21.54  \\
ChemLLM-7B-chat&17.52& 3.42 & 13.45   & 6.02 &  19.16  & 8.81 &  15.73  & 8.15 &  18.01  & 13.45  \\
\midrule

\multicolumn{1}{c}{}                        & \multicolumn{1}{c}{}       & \multicolumn{1}{c}{}     & \multicolumn{1}{c}{}       & \multicolumn{1}{c}{}     & \multicolumn{1}{c}{}       & \multicolumn{1}{c}{}     & \multicolumn{1}{c}{}       & \multicolumn{1}{c}{}     & \multicolumn{1}{c}{}       & \multicolumn{1}{c}{}    \\ \hline
\multicolumn{1}{c|}{\multirow{2}{*}{Model}} & \multicolumn{2}{c|}{\textbf{PriKG}}                 & \multicolumn{2}{c|}{\textbf{HipKG}}                  & \multicolumn{2}{c|}{\textbf{GoKG}}          & \multicolumn{2}{c|}{\textbf{BioText}}           & \multicolumn{2}{c}{\textbf{MatText}}                \\ \cline{2-11} 
\multicolumn{1}{c|}{}                       & \multicolumn{1}{c}{Direct} & \multicolumn{1}{c|}{Context} & \multicolumn{1}{c}{Direct} & \multicolumn{1}{c|}{Context} & \multicolumn{1}{c}{Direct} & \multicolumn{1}{c|}{Context} & \multicolumn{1}{c}{Direct} & \multicolumn{1}{c|}{Context} & \multicolumn{1}{c}{Direct} & \multicolumn{1}{c}{Context} \\ \hline

GPT-4o&17.44&{{54.80}}&14.42&68.50&43.47&74.32&53.41&79.03&41.28&64.57\\
GPT-4o-mini&16.48&52.64&10.99&65.20&42.80&73.14&{55.68}&{{79.24}}&{48.09}&{{65.00}}\\
claude-3.5-sonnet&{26.80}&45.96&{21.55}&{{75.78}}&{{45.59}}&{{84.07}}&55.68&58.06&41.60&  61.49\\
\midrule
DeepSeek-R1 & 10.29  & {{58.20}}  & 10.69  &  69.66  &  31.40   &   {79.18} & 56.43   &  {74.79}  &   45.30 & {63.09}   \\

Qwen3-8B & 12.80  & 54.48  & 10.82 &   {74.68} & 31.44   &  73.98  & 48.76   &   69.73 &   39.04 &  55.11 \\

Deepsee-V3&{17.33}& {51.92} &  14.76  & {63.42} &  {39.75}  & {72.29} &  60.07  & 66.74 &  {51.18}  &  45.31 \\

Llama4-Maverick & 20.18  &  49.28  & 17.85 &  64.81 &  {43.81}   &   72.71 & {{61.77}}   &  63.02  &   {{54.79}} & 54.15  \\
Llama4-Scout & {23.84} & 48.08   & {{21.97}}&  59.57 &   37.54  &   66.27 &  55.58 &  61.88  &  42.98  & 48.51  \\

Llama3.1-70B-it&14.40& 44.00 &{15.88} & 59.31 & 32.12  & 70.17 &  49.80& 66.53 & 40.21  &  51.91 \\
Qwen2.5-7B-it&16.56&45.60&9.87&53.99&33.64&62.46&47.11& {68.18} &36.81& {59.68} \\
GLM4-9B-Chat&16.72&44.56&  11.93  &57.94& 30.17 & 60.51 &  47.52  & 67.77 & 36.38   &  50.96 \\

Llama3.1-8B-it&16.24&46.56&14.51&52.62&35.51&59.32&47.31& 64.26    &37.34&49.36\\

Gemma2-9B-it&15.36&50.48&  9.96  &56.57&  31.27  & 57.29&  51.81  & 37.77& 35.88  & 29.67\\
Ministral-8B-it&15.05& 37.92 &13.24& 48.51 &28.27& 52.88 &41.84& 48.14 &32.23&45.32 \\

\midrule
ChemDFM-v1.5-8B&\textbf{{33.66}}& {30.72} &{{30.21}}& {49.70} &{39.84}& {56.44} &{50.88}& 26.11 &{30.83}&  18.91 \\
SciGLM-6B&15.20& 20.88 &18.80& 21.46 &25.93& 28.31 &33.44& {44.17} &21.63&{ 31.35}  \\
LlaSMol-Mistral-7B&15.52& 19.84 &20.17& 22.83 &23.39& 29.92 &33.85&  33.13&23.58& 20.98  \\
ChemLLM-7B-chat&16.80 & 5.92 & 23.86 & 5.15 &  27.80  & 15.51 &  45.92 & 39.94 &  30.44  & 22.67  \\

\bottomrule

\end{tabular}%
}

\label{tab:without-RAG}
\end{table*}

\section{Prompts for Data Generation}\label{ap:construct_prompt}

We present distinct prompt templates for each of the four capabilities below. 
\begin{itemize}[left=0pt]
    \item  \noindent \textbf{A prompt for generating questions about relevant information identification}
\begin{tcolorbox}[colback=green!3, colframe=black, arc=2mm,left=3pt,right=3pt,
    boxsep=5pt,    boxrule=0.5pt,    breakable]  
    
    \textbf{System Message:} \\ 
    You're a brilliant in scientific domain.\\
    
    \textbf{User Message:} \\
    You will be provided with several triples from PriKG that form a path connecting a starting point to an endpoint. Based on this path, you need to generate a scientific question designed to test the respondent's ability to find the correct answer in the noise, with information from the knowledge graph. The question types can be Q\&A or fill-in-the-blank. The answers to QA questions should be simple, concise, and easily verifiable phrases, not long sentences. \\

    Start Node: \{start$\_$node\} \\
    End Node: \{end$\_$node\}  \\
    Path: \{data['path']\} \\

    Triples: \\
    \{data['triplets']\} \\

    Please generate a scientific question based on this information. Ensure that the question requires the respondent to find the correct answer in the noise in the knowledge graph and the difficulty level should be moderate. Please output the question in JSON format only. Do not output anything other than the JSON format. The JSON format should look like this: \\

    \{ \\
    "question$\_$type": "[Question type]", \\
    "question": "[Question or rejection]", \\
    "answer": "[Answer]" \\
    \} \\
\end{tcolorbox}
   
    \item  \noindent \textbf{A prompt for generating questions about information-absence detection}
\begin{tcolorbox}[colback=green!3, colframe=black, arc=2mm,left=3pt,right=3pt,
    boxsep=5pt,    boxrule=0.5pt,    breakable]  
    
    \textbf{System Message:} \\ 
    You're  brilliant in the scientific domain.\\
    
    \textbf{User Message:} \\
    You will be provided with a relevant information identification question and its corresponding correct context, also context. Your task is to remove the correct contextual information from the context. \\

    Do not alter the form of the question. Output the question in JSON format only, without any additional text. The JSON format should adhere to the following structure: \\
    \{ \\
    "question": "[Question or rejection]", \\
    "answer": "[Here is the answer]" \\
    \} \\
    
    Next is the context you need to use: \{Contexts\}
\end{tcolorbox}

    \item  \noindent \textbf{A prompt for generating questions about multi-source information integration}
\begin{tcolorbox}[colback=green!3, colframe=black, arc=2mm,left=3pt,right=3pt,
    boxsep=5pt,    boxrule=0.5pt,    breakable]  
    \textbf{System Message:} \\ 
    You're brilliant in the scientific domain.\\
    \textbf{User Message:} \\
    You will be provided with several data entries describing various properties of different materials. Based on these properties, you need to generate a scientific question that tests the respondent's ability to retrieve, integrate, and analyze information from the table. \\
    Please follow the instructions below to generate the question and answer: \\
    1. The question should be in Q\&A format, starting with sentence like "Given the following four materials: mp-xxxxx, mp-xxxxx, mp-xxxxx, mp-xxxxx" or "Which of the following materials, mp-xxxxx, mp-xxxxx, mp-xxxxx, mp-xxxxx". \\
    2. The question should focus on a single numeric property of the materials that is representative of the material and comparable. \\
    3. The question should involve comparing the values of this property and identifying the result. \\
    4. The answer should be the material ID of the material with the correct value, and the answer must be one of the materials listed in the question. \\
    Please output the question in JSON format only. Do not output anything
    other than the JSON format. The JSON format should look like this: \\
    \{ \\
    "question": "[Question or rejection]", \\
    "answer": "[Answer]" \\
    \} \\
    Next is the data entries you need to use:
    \vspace{-0.5cm}
        \begin{flushleft} 
            \renewcommand{\arraystretch}{1.5}  
            \setlength{\arrayrulewidth}{1pt}  
            \resizebox{\linewidth}{!}{  
            \begin{tabular}{|c|}
                \hline
                \rowcolor{gray!20} \multicolumn{1}{|l|}{\textbf{Material ID, Formula ... Sites ... Volume, Density}} \\  
                \hline
                mp-xxxxx ......................................................... \\ 
                \hline
                mp-xxxxx ......................................................... \\ 
                \hline
                mp-xxxxx ......................................................... 
                \\ 
                \hline
                mp-xxxxx ......................................................... \\ 
                \hline
            \end{tabular}
            }
        \end{flushleft}
    
\end{tcolorbox}

    \item  \noindent \textbf{A prompt for generating questions about reasoning}
\begin{tcolorbox}[colback=green!3, colframe=black, arc=2mm, left=3pt,right=3pt,
    boxsep=5pt,    boxrule=0.5pt,   breakable]  
    \textbf{System Message:} \\ 
    You're brilliant in the scientific domain.\\
    \textbf{User Message:} \\
    Please write a scientific reasoning question based on the following article. Treat the paper as consisting of two parts. The first part includes the introduction, background, methods, and experimental results. The second part contains the conclusions and analysis derived from the first part. The goal of the question is to test the ability to infer the second part based on the summary of the first part, without knowing the premises of the first part. Therefore, the question should be based on the first part. \\
    Please follow the instructions below to generate the question and answer:
    1. The question should be a multiple-choice question with four options, one or more of which is correct, and the others are incorrect. \\
    2. The difficulty level of the question is high and should involve summarizing, generalizing, and reasoning, rather than simple information retrieval or verification. The question should require at least a university-level education to answer. \\
    3. The answer to the question should not be directly available from the first part paragraphs. It should not be directly deducible but should require complex reasoning to arrive at the correct answer. \\
    4. Incorrect options should contain errors or deviations from the original content. The incorrect options should sound reasonable, but the content must be wrong. \\
    5. If you feel you cannot generate a question or are uncertain about the correctness of the question, please output “[Unable to generate question]”. \\
    6. The question should be very difficult. If you feel you cannot provide a high-difficulty question, please output “[Unable to generate question]”. \\
    Please output the question in JSON format only. Do not output anything other than the JSON format. The JSON format should look like this:   \\
    \{
    "question": "[Question or rejection]", \\
    "options": \{   \\
        "A": "[Option A]",  \\
        "B": "[Option B]",   \\
        "C": "[Option C]",    \\
        "D": "[Option D]"     \\
    \}, \\
    "answer": "[A or B or C or D]"  \\
    \}  \\
    Next is the full text of the article:\\
    \{Papers\}
    
\end{tcolorbox}

\end{itemize}

\section{Data Quality Verification}\label{ap:data_quality}

\noindent\textbf{LLM as a Judge:} 
We use advanced LLMs (e.g., GPT-4o) as automated evaluators to verify that each generated answer is both extractable and logically deducible from the relevant context, ensuring factual consistency and relevance. The prompt is presented below.

\begin{tcolorbox}[colback=green!3, colframe=black, arc=2mm,left=3pt,right=3pt,
    boxsep=5pt,    boxrule=0.5pt,    breakable]
    \textbf{System Message:} \\
    You're a highly capable evaluator in the scientific domain. \\
    \textbf{User Message:} \\
    Below is a question, its relevant context, and an answer. Your task is to verify whether the answer meets the following standard: \\
    1. The answer must be explicitly extractable or logically deducible from the provided context. \\
    2. The answer must adhere strictly to the relevant information in the context and be factually correct. \\
    3. If the answer meets the standard, output "Yes". If it does not meet the standard, output "No". \\
    \textbf{[Relevant Context start]} \\
    \{Context\} \\
    \textbf{[Relevant Context end]} \\
    
    \textbf{[Question start]} \\
    \{Question\} \\
    \textbf{[Question end]} \\
    
    \textbf{[Answer start]} \\
    \{Answer\} \\
    \textbf{[Answer end]} \\
    Please evaluate and output either "Yes" or "No" based on the above criteria.
\end{tcolorbox}

\noindent\textbf{Human Expert Evaluation:}  To further ensure the quality and accuracy of the generated data, we subjected the data that passed the initial LLM validation to manual review by five PhD-level researchers with strong STEM backgrounds. These experts were tasked with thoroughly evaluating each instance based on the following three criteria:  
\begin{enumerate}
    \item Whether the question effectively tests the intended competency, ensuring that it is aligned with the targeted skill or knowledge domain and accurately reflects the underlying construct it aims to assess.
    \item Whether the question is expressed clearly and logically, such that its wording is unambiguous, coherent, and easily understood by both human evaluators and automated systems, thereby minimizing potential misinterpretations
    \item Whether the given contexts fully support the given answer and is factually correct, which requires that the answer not only directly derives from or can be logically inferred based on the supporting materials, but also adheres to facts and scientific evidence. Together, these criteria are designed to ensure the evaluation process's validity, clarity, and reliability.
\end{enumerate}
Only instances that received "Yes" for all three criteria were accepted. After the experts reviewed all instances, the results revealed that 90.83\% of the instances met the required high-quality standards.

We invited five PhD-level researchers with STEM backgrounds, including two domain experts in bioinformatics.
We compensated them based on the number of questions reviewed. We paid \$30 for every 100 questions reviewed, totaling \$3,300 for 11k questions. The entire review process took one week.

\section{Dataset Examples}\label{ap:QA_examples}
In this part, we demonstrate several examples of questions aligned with four core competencies. For each competency, we present three examples corresponding to three distinct data modalities. \\

\noindent\textbf{(1) Relevant Information Identification}

\begin{tcolorbox}[colback=blue!3, colframe=black, arc=2mm, breakable, left=3pt,right=3pt,
    boxsep=5pt,    boxrule=0.5pt,     colframe=black!70,  fonttitle=\bfseries, title = Unstructured Text]

    \textbf{System Message:} \\ Please answer the scientific questions based on the content. Your answer only needs to include the one or more correct option labels, not the full options. You should give your answer directly  without any other characters. \\

    \textbf{User Message:} \\
     What is the primary objective of the statistical framework proposed in the paper 'Augmented Doubly Robust Post-Imputation Inference for Proteomic Data'? \\
    (A) To develop a method for directly measuring protein abundances without missing values. \\
    (B) To create a statistical framework that offers valid and efficient inference for proteomic data by addressing the challenge of missing values. \\
    (C) To replace the Plugin method with a simpler imputation strategy that discards missing values. \\
    (D) To develop a tool that solely relies on low-dimensional covariates for analyzing proteomic data.
    \vspace{-0.2cm}
    
    \begin{flushleft}
        \renewcommand{\arraystretch}{1.5}
        \setlength{\arrayrulewidth}{1pt}
    
        {\textbf{Corpus 1 ......... (Irrelevant Content)} } \\[0.5em]
        {\textbf{Corpus 2 ......... (Irrelevant Content)} } \\[0.5em]
        {\textbf{Corpus 3 ......... (Correct Content)} } \\[0.5em]
        {\textbf{Corpus 4 ......... (Irrelevant Content)} } \\[0.5em]
        {\textbf{Corpus 5 ......... (Irrelevant Content)} }

    \end{flushleft}

    \textbf{Expected Answer:} \\ B
\end{tcolorbox}

\begin{tcolorbox}[colback=blue!3, colframe=black, arc=2mm, breakable, left=3pt,right=3pt,
    boxsep=5pt,    boxrule=0.5pt,     colframe=black!70,  fonttitle=\bfseries, title = Table] %
    \textbf{System Message:} \\ Please answer the scientific questions based on the content. You should give your answer directly without any other characters. \\

    \textbf{User Message:} \\
    For the material with CID 13182, what is its inchikey?
    
    \vspace{-0.4cm}
    {
        \begin{flushleft}  
            \renewcommand{\arraystretch}{1.5}  
            \setlength{\arrayrulewidth}{1pt}  
            \resizebox{\linewidth}{!}{  
            \begin{tabular}{|c|}
                \hline
                \rowcolor{gray!20} \multicolumn{1}{|l|}{\textbf{cid, mw, mf, xlog... inchikey ... exactmass}} \\  
                \hline
                CID XXXXX ....................................................... × \\ 
                \hline
                \rowcolor{green!20} CID 13182 ....................................................... $\checkmark$ \\ 
                \hline
                CID XXXXX ....................................................... × \\ 
                \hline
                CID XXXXX ....................................................... × \\ 
                \hline
            \end{tabular}
            }
        \end{flushleft}
    }

    \textbf{Expected Answer:} \\ ARBSJUHHKXRHAD-UHFFFAOYSA-N
\end{tcolorbox}

\begin{tcolorbox}[colback=blue!3, colframe=black, arc=2mm, breakable, left=3pt,right=3pt,
    boxsep=5pt,    boxrule=0.5pt,     colframe=black!70,  fonttitle=\bfseries, title = Knowledge Graph]     %
    
    \textbf{System Message:} \\ Please answer the scientific questions based on the content. You should give your answer directly without any other characters. \\

    \textbf{User Message:} \\
    How is the gene or protein known as 'GDPD3' connected to the anatomical structure called the 'lymph node'?
    
    {
        \begin{flushleft}  
            \renewcommand{\arraystretch}{1.5}  
            \setlength{\arrayrulewidth}{1pt}  
            \resizebox{\linewidth}{!}{  
            \begin{tabular}{|c|}
                \hline
                \rowcolor{gray!20} \multicolumn{1}{|c|}{\textbf{[x$\_$name, \quad  relation,   \quad y$\_$name]}} \\  
                \hline
                [Stiripentol,   \quad  drug$\_$drug,  \quad   Sumatriptan ]  ×\\ 
                \hline
                \rowcolor{green!20} {[GDPD3,  anatomy$\_$protein$\_$present, lymph node]} $\checkmark$ \\ 
                \hline
                {[TROAP,\quad protein$\_$protein, \quad  NBPF19]} × \\ 
                \hline
                [DB00351, \quad drug$\_$drug,\quad Reserpine ] × \\ 
                \hline
            \end{tabular}
            }
        \end{flushleft}
    }

    \textbf{Expected Answer:} \\ anatomy$\_$protein$\_$present
\end{tcolorbox}

\noindent \textbf{(2) Information-absence Detection}

\begin{tcolorbox}[colback=blue!3, colframe=black, arc=2mm, breakable, left=3pt,right=3pt,
    boxsep=5pt,    boxrule=0.5pt,     colframe=black!70,  fonttitle=\bfseries, title = Unstructured Text]

    \textbf{System Message:} \\ Please answer the scientific questions based on the content. Your answer only needs to include the one or more correct option labels, not the full options. You should give your answer directly without any other characters. \\

    \textbf{User Message:} \\
    What key feature of elliptically geared isostatic metamaterials enables their nonlinear topological transitions? \\
    (A) The unique soliton-induced mechanical deformation in linear gear mechanisms. \\
    (B) The nonlinear Berry phase transition facilitated by geometric nonlinearity. \\
    (C) The presence of circular gear geometry that allows reversible elastic deformation. \\
    (D) The linear topological index change due to minor gear rotations.
    \vspace{-0.2cm}
    
    \begin{flushleft}
        \renewcommand{\arraystretch}{1.5}
        \setlength{\arrayrulewidth}{1pt}
    
        {\textbf{Corpus 1 ......... (Irrelevant Content)} } \\[0.5em]
        {\textbf{Corpus 2 ......... (Irrelevant Content)} } \\[0.5em]
        {\textbf{Corpus 3 ......... (Irrelevant Content)} } \\[0.5em]
        {\textbf{Corpus 4 ......... (Irrelevant Content)} } \\[0.5em]
        {\textbf{Corpus 5 ......... (Irrelevant Content)} }

    \end{flushleft}

    \textbf{Expected Answer:} \\ I cannot answer the question due to insufficient information in the retrieved data.
\end{tcolorbox}

\begin{tcolorbox}[colback=blue!3, colframe=black, arc=2mm, breakable, left=3pt,right=3pt,
    boxsep=5pt,    boxrule=0.5pt,     colframe=black!70,  fonttitle=\bfseries, title = Table]

    \textbf{System Message:} \\ Please answer the scientific questions based on the content.  You should give your answer directly without any other characters. \\
   
    \textbf{User Message:} \\
    For the material with ID mp-768851, what is its number of site?
    
    \vspace{-0.4cm}
    {
        \begin{flushleft} 
            \renewcommand{\arraystretch}{1.5}  
            \setlength{\arrayrulewidth}{1pt}  
            \resizebox{\linewidth}{!}{  
            \begin{tabular}{|c|}
                \hline
                \rowcolor{gray!20} \multicolumn{1}{|l|}{\textbf{Material ID, Formula ... Sites ... Volume, Density}} \\  
                \hline
                mp-xxxxx ....................................................... × \\ 
                \hline
                mp-xxxxx ....................................................... × \\ 
                \hline
                mp-xxxxx ....................................................... × \\ 
                \hline
                mp-xxxxx ....................................................... × \\ 
                \hline
            \end{tabular}
            }
        \end{flushleft}
    }

    \textbf{Expected Answer:} \\ I cannot answer the question due to insufficient information in the retrieved data.
\end{tcolorbox}

\begin{tcolorbox}[colback=blue!3, colframe=black, arc=2mm, breakable, left=3pt,right=3pt,
    boxsep=5pt,    boxrule=0.5pt,     colframe=black!70,  fonttitle=\bfseries, title = Knowledge Graph]

    \textbf{System Message:} \\ Please answer the scientific questions based on the content.  You should give your answer directly without any other characters. \\

    \textbf{User Message:} \\
    How are the genes "nbc 1" and "nbc 3" related?

    \vspace{0.0cm}
    {
        \begin{flushleft}  
            \renewcommand{\arraystretch}{1.5}  
            \setlength{\arrayrulewidth}{1pt}  
            \resizebox{\linewidth}{!}{  
            \begin{tabular}{|c|}
                \hline
                \rowcolor{gray!20} \multicolumn{1}{|c|}{\textbf{[x$\_$name, \quad  relation,   \quad y$\_$name]}} \\  
                \hline
                [Stiripentol,   \quad  drug$\_$drug,  \quad   Sumatriptan ]  ×\\ 
                \hline
               {[GDPD3,  anatomy$\_$protein$\_$present, lymph node]} × \\ 
                \hline
                {[TROAP,\quad protein$\_$protein, \quad  NBPF19]} × \\ 
                \hline
                [DB00351, \quad drug$\_$drug,\quad Reserpine ] × \\ 
                \hline
            \end{tabular}
            }
        \end{flushleft}
    }

    \textbf{Expected Answer:} \\ I cannot answer the question due to insufficient information in the retrieved data.
\end{tcolorbox}

\noindent\textbf{(3) Multi-source Information Integration}

\begin{tcolorbox}[colback=blue!3, colframe=black, arc=2mm, breakable, left=3pt,right=3pt,
    boxsep=5pt,    boxrule=0.5pt,     colframe=black!70,  fonttitle=\bfseries, title = Unstructured Text]

    \textbf{System Message:} \\ Please answer the scientific questions based on the content. Your answer only needs to include the one or more correct option labels, not the full options. You should give your answer directly  without any other characters. \\

    \textbf{User Message:} \\
    Based on the findings of the study, what is the primary long-term effect of local SBRT/IL-12 therapy on the bone marrow of treated mice? \\
    (A) A permanent increase in hematopoietic stem cells (HSCs). \\
    (B) A transient increase in IL-12 levels followed by long-term activation of myeloid cells. \\
    (C) A significant reduction in hematopoietic stem cells (HSCs) accompanied by skewing toward a myeloid lineage bias. \\
    (D) A substantial increase in IL-12 and IFN$\gamma$ concentrations in the bone marrow.
    \vspace{-0.2cm}
    
    \begin{flushleft}
        \renewcommand{\arraystretch}{1.5}
        \setlength{\arrayrulewidth}{1pt}
    
        {\textbf{Corpus 1 ......... (Irrelevant Content)} } \\[0.5em]
        {\textbf{Corpus 2 ......... (Correct Content)} } \\[0.5em]
        {\textbf{Corpus 3 ......... (Correct Context)} } \\[0.5em]
        {\textbf{Corpus 4 ......... (Irrelevant Content)} } \\[0.5em]
        {\textbf{Corpus 5 ......... (Irrelevant Content)} }

    \end{flushleft}

    \textbf{Expected Answer:} \\ C
\end{tcolorbox}

\begin{tcolorbox}[colback=blue!3, colframe=black, arc=2mm, breakable, left=3pt,right=3pt,
    boxsep=5pt,    boxrule=0.5pt,     colframe=black!70,  fonttitle=\bfseries, title = Table]

    \textbf{System Message:} \\ Please answer the scientific questions based on the content. You should give your answer directly without any other characters. \\
  
    \textbf{User Message:} \\
    Given the following isotopes ID: NDS-54874, NDS-30453, NDS-69167, NDS-58315, tell me which isotopes has the largest energy?
    
    \vspace{-0.4cm}
    {
        \begin{flushleft}  
            \renewcommand{\arraystretch}{1.5}  
            \setlength{\arrayrulewidth}{1pt}  
            \resizebox{\linewidth}{!}{  
            \begin{tabular}{|c|}
                \hline
                \rowcolor{gray!20} \multicolumn{1}{|l|}{\textbf{id, Z, N, symbol... energy[kev]... relative intensity}} \\  
                \hline
                NDS-XXXXX ....................................................... × \\ 
                \hline
                \rowcolor{green!20} NDS-30453 ....................................................... $\checkmark$ \\ 
                \hline
                \rowcolor{green!20} NDS-58315 ....................................................... $\checkmark$ \\ 
                \hline
                NDS-XXXXX ....................................................... × \\ 
                \hline
                \rowcolor{green!20} NDS-69167....................................................... $\checkmark$ \\ 
                \hline
                NDS-XXXXX ....................................................... × \\ 
                \hline
                \rowcolor{green!20} NDS-54874....................................................... $\checkmark$ \\ 
                \hline
            \end{tabular}
            }
        \end{flushleft}
    }

    \textbf{Expected Answer:} \\ NDS-69167
\end{tcolorbox}

\begin{tcolorbox}[colback=blue!3, colframe=black, arc=2mm, breakable, left=3pt,right=3pt,
    boxsep=5pt,    boxrule=0.5pt,     colframe=black!70,  fonttitle=\bfseries, title = Knowledge Graph]

    \textbf{System Message:} \\ Please answer the scientific questions based on the content. You should give your answer directly without any other characters. \\

    \textbf{User Message:} \\
    Could you list the substances that have the potential to interact with DB131$\_$HUMAN?

    {
        \begin{flushleft}  
            \renewcommand{\arraystretch}{1.5}  
            \setlength{\arrayrulewidth}{1pt}  
            \resizebox{\linewidth}{!}{  
            \begin{tabular}{|c|}  
                \hline  
                
                \rowcolor{green!20} {[DB131$\_$HUMAN,\quad Confidence: 0.63, \quad LRC8A$\_$HUMAN] }$\checkmark$ \\ 
                \hline
                [ATX1$\_$HUMAN,\quad Confidence: 0.49, \quad PK3CA$\_$HUMAN] × \\
                \hline

                \rowcolor{green!20} {[DB131$\_$HUMAN,\quad Confidence: 0.63, \quad RBM12$\_$HUMAN] }$\checkmark$ \\ 

                \hline
                [RASN$\_$HUMAN,\quad Confidence: 0.73, \quad PVRL3$\_$HUMAN] × \\
                \hline
                \rowcolor{green!20} {[DB131$\_$HUMAN,\quad Confidence: 0.65, \quad AHNK2$\_$HUMAN] }$\checkmark$ \\ 

                \hline
            \end{tabular}
            }
        \end{flushleft}
    }

    \textbf{Expected Answer:} \\ "LRC8A$\_$HUMAN", \\
            "AHNK2$\_$HUMAN", \\
            "RBM12$\_$HUMAN"
\end{tcolorbox}

\textbf{(4) Context-aware Inference,}

\begin{tcolorbox}[colback=blue!3, colframe=black, arc=2mm, breakable, left=3pt,right=3pt,
    boxsep=5pt,    boxrule=0.5pt,     colframe=black!70,  fonttitle=\bfseries, title = Unstructured Text]

    \textbf{System Message:} \\ Please answer the scientific questions based on the content. Your answer only needs to include the one or more correct option labels, not the full options. You should give your answer directly  without any other characters. \\

    \textbf{User Message:} \\
    Based on the methods and results described in the first part of the study on epitaxial growth of GaAs on Si(001), which of the following is the most plausible reasoning for the effectiveness of the GaSb buffer layer in reducing defect densities such as threading dislocations and antiphase boundaries in the GaAs layer? \\
    (A) The antimonides, such as GaSb, have a significant lattice mismatch with silicon, leading to the generation of interfacial misfit dislocation arrays that efficiently alleviate strain without forming threading dislocations. \\
    (B) The presence of the GaSb buffer layer increases the thickness of the overall film, which inherently reduces the formation of threading dislocations and antiphase boundaries in the GaAs layer. \\
    (C) The GaSb buffer layer chemically reacts with silicon to form a new compound at the interface, which serves as an ideal seed layer for epitaxial GaAs growth, minimizing defect densities. \\
    (D) The GaSb buffer layer promotes planar defects, such as stacking faults, that counterbalance and neutralize threading dislocations and antiphase boundaries in the GaAs layer.
    \vspace{-0.2cm}
    
    \begin{flushleft}
        \renewcommand{\arraystretch}{1.5}
        \setlength{\arrayrulewidth}{1pt}
    
        {\textbf{Corpus 1 ......... (Irrelevant Content)} } \\[0.5em]
        {\textbf{Corpus 2 ......... (Correct Content)} } \\[0.5em]
        {\textbf{Corpus 3 ......... (Irrelevant Content)} } \\[0.5em]
        {\textbf{Corpus 4 ......... (Irrelevant Content)} } \\[0.5em]
        {\textbf{Corpus 5 ......... (Irrelevant Content)} }

    \end{flushleft}

    \textbf{Expected Answer:} \\ A
\end{tcolorbox}

\begin{tcolorbox}[colback=blue!3, colframe=black, arc=2mm, breakable, left=3pt,right=3pt,
    boxsep=5pt,    boxrule=0.5pt,     colframe=black!70,  fonttitle=\bfseries, title = Table]

    \textbf{System Message:} \\ Please answer the scientific questions based on the content. Your answer only needs to include the one or more correct option labels, not the full options. You should give your answer directly  without any other characters. \\

    \textbf{User Message:} \\
    Comparing materials mp-760154 and mp-1208151, which statement is correct? \\
    (A) Both materials have identical band gaps and belong to the same crystal system.\\
    (B) The material mp-1208151 has a much larger volume and higher density than mp-760154.\\
    (C) The material mp-760154 is metallic, while mp-1208151 is semiconducting.\\
    (D) Both materials are predicted to be stable with similar formation energies.
    
    \vspace{-0.4cm}
    {
        \begin{flushleft}  
            \renewcommand{\arraystretch}{1.5}  
            \setlength{\arrayrulewidth}{1pt}  
            \resizebox{\linewidth}{!}{  
            \begin{tabular}{|c|}
                \hline
                \rowcolor{gray!20} \multicolumn{1}{|l|}{\textbf{Material ID, Formula ... Sites ... Volume, Density}} \\  
                \hline
                mp-xxxxx ....................................................... × \\ 
                \hline
                \rowcolor{green!20} mp-760154 ..................................................... $\checkmark$ \\ 
                \hline
                mp-xxxxx ....................................................... × \\ 
                \hline
                \rowcolor{green!20} mp-1208151 ..................................................... $\checkmark$ \\ 
                \hline
                mp-xxxxx ....................................................... × \\ 
                \hline
            \end{tabular}
            }
        \end{flushleft}
    }

    \textbf{Expected Answer:} \\ B
\end{tcolorbox}
  
\begin{tcolorbox}[colback=blue!3, colframe=black, arc=2mm, breakable, left=3pt,right=3pt,
    boxsep=5pt,    boxrule=0.5pt,     colframe=black!70,  fonttitle=\bfseries, title = Knowledge Graph]

    \textbf{System Message:} \\ Please answer the scientific questions based on the content. You should give your answer directly without any other characters. \\

    \textbf{User Message:} \\
    Given that there exists a shared intermediate term, fill in the blank: GO:0003399 (cytoneme morphogenesis)  $\_$$\_$$\_$$\_$$\_$ GO:0048858 (cell projection morphogenesis).
    
    {
        \begin{flushleft}  
            \renewcommand{\arraystretch}{1.2}  
            \setlength{\arrayrulewidth}{1pt}
            \small
            \resizebox{\linewidth}{!}{  
            \begin{tabular}{|c|}  
                \hline  
                \rowcolor{green!20} {[GO:0003399,\quad is$\_$a, \quad GO:0120039 ]}$\checkmark$ \\ 
                \hline
                [GO:0086014,\quad namespace, \quad biological$\_$process ] × \\
                \hline
                [GO:0003399,\quad namespace, \quad biological$\_$process ] × \\
                \hline
                \rowcolor{green!20} {[GO:0120039,\quad is$\_$a, \quad GO:0048858 ]}$\checkmark$ \\ 
                \hline
                [GO:0048686,\quad is$\_$a, \quad 0022603 ] × \\
                \hline
            \end{tabular}
            }
        \end{flushleft}
    }
    \vspace{0.2cm}

    \textbf{Expected Answer:} \\ is$\_$a
\end{tcolorbox}

\section{Detailed Data Source}\label{ap:data_source}
Table \ref{tab:data_source} provides detailed information on all databases we used to construct our \OURS{}, including their URL, description, and license.
\begin{table*}[htbp]
    \small
    \centering
    \caption{Detailed URL, description, and license of the source data involved in this paper.}
    \renewcommand{\arraystretch}{1.5} 
    \resizebox{\linewidth}{!}{ 
        \begin{tabular}{p{1.5cm}p{5cm}p{8cm}p{2cm}}
            \toprule
            \textbf{Dataset Name} & \textbf{URL} & \textbf{Database Description} & \textbf{License} \\
            \midrule
            MatText & \href{https://arxiv.org/}{arxiv.org} & A compilation of material domain research publications. & CC BY\\
            BioText & \href{https://bio-protocol.org/}{bio-protocol.org} & A peer-reviewed, open-access journal publishing step-by-step life science protocols. & CC BY 4.0 \\
            MatTab & \href{https://next-gen.materialsproject.org}{next-gen.materialsproject.org} & Offer data on over 160,000 inorganic compounds, like crystal structures. & CC BY 4.0 \\
            IaeaTab & \href{https://www-nds.iaea.org/relnsd/NdsEnsdf/QueryForm.html}{www-nds.iaea.org} & Provide data on evaluated nuclear structure and decay data, including energy levels. & CC BY-NC \\
            ProtTab & \href{https://pubchem.ncbi.nlm.nih.gov/classification/\#hid=72}{pubchem.ncbi.nlm.nih.gov-protein} & Offer chemical property information of more than 320,000 common compounds. & CC BY\\
            MolTab & \href{https://pubchem.ncbi.nlm.nih.gov/classification/\#hid=72}{pubchem.ncbi.nlm.nih.gov-chemical} & Offer protein information of more than 60,000 common proteins. & CC BY \\
            GoKG & \href{https://geneontology.org/docs/download-ontology/}{geneontology.org} & A standardized framework for biological knowledge, covering molecular function, cellular component, and biological process. & CC BY 4.0 \\
            HipKG & \href{https://cbdm-01.zdv.uni-mainz.de/\~mschaefer/hippie/}{cbdm-01.zdv.uni-mainz.de} & Offer confidence scored and functionally annotated human protein-protein interactions.  & CC BY 4.0 \\
            PhaKG & \href{https://zenodo.org/records/4077338}{zenodo.org/records} & A biomedical KG comprising over 500,000 interconnections between genes, drugs, etc. & CC BY-NC 4.0 \\
            PriKG & \href{https://dataverse.harvard.edu/dataset.xhtml?persistentId=doi:10.7910/DVN/IXA7BM}{dataverse.harvard.edu} & A KG integrating 20 biomedical resources to describe over 17,000 diseases and 4,000,000 relationships across ten biological scales. & MIT License\\
            \bottomrule
        \end{tabular}
    }
    
    \label{tab:data_source}
\end{table*}

\section{Detailed Model Descriptions}\label{ap:models}
We have selected 18 high-performing LLMs with different scales for this paper. LLama3.1-70b-it,  Llama4-Scout and Llama4-Maverick are accessed via the NVIDIA NIM APIs. DeepSeek-R1, DeepSeek-V3, and proprietary models are accessed via their official APIs. The remaining open-source models are deployed locally on a server equipped with two NVIDIA GeForce RTX 3090 GPUs. The detailed information of these models is shown in Table \ref{tab:model_description}.

\begin{table*}[htbp!]
\small
\caption{Overview of the LLMs assessed in our experimental framework.}
\renewcommand{\arraystretch}{1.3} 
\resizebox{\textwidth}{!}{%
\begin{tabular}{p{3.0cm} p{2.5cm} p{2.0cm} p{2.0cm} p{1.5cm} p{6.5cm}}
\toprule
\textbf{Model Name} & \textbf{Creator} & \textbf{Domain} & \textbf{\#Parameters} & \textbf{Access} & \textbf{URL} \\ 
\midrule
GPT-4o & OpenAI & General & undisclosed & Official API & \href{https://chat.openai.com}{https://chat.openai.com} \\
GPT-4o-mini & OpenAI & General & undisclosed & Official API & \href{https://chat.openai.com}{https://chat.openai.com} \\
Claude-3.5-Sonnet & Anthropic & General & undisclosed & Official API & \href{https://claude.ai}{https://claude.ai} \\
\midrule
DeepSeek-V3 & DeepSeek & General & 671B & Official API & \href{https://www.deepseek.com}{https://www.deepseek.com}\\

DeepSeek-R1 & DeepSeek & General & 671B & Official API & \href{https://www.deepseek.com}{https://www.deepseek.com}\\
Llama3.1-70B-it & Meta & General & 70B & NVIDIA NIM API & \href{https://llama.meta.com/llama3}{https://llama.meta.com/llama3} \\
Llama3.1-8B-it & Meta & General & 8B & Weights & \href{https://llama.meta.com/llama3}{https://llama.meta.com/llama3} \\
Llama4-Maverick& Meta & General &  400B(17B×128 Experts)  & NVIDIA NIM API & \href{https://www.llama.com/models/llama-4/}{https://www.llama.com/models/llama-4/} \\
Llama4-Scout& Meta & General &  109B(17B×16 Experts)   & NVIDIA NIM API & \href{https://www.llama.com/models/llama-4/}{https://www.llama.com/models/llama-4/} \\
Qwen2.5-7B-it & Alibaba & General & 7B & Weights & \href{https://qwenlm.github.io/}{https://qwenlm.github.io/} \\
Qwen3-8B & Alibaba & General & 8B & Weights & \href{https://qwenlm.github.io/}{https://qwenlm.github.io/} \\

GLM4-9B-Chat & Tsinghua\&Zhipu & General & 9B & Weights & \href{https://huggingface.co/THUDM/glm-4-9b-chat}{https://huggingface.co/THUDM/glm-4-9b-chat} \\
Gemma2-9B-it & Google & General & 9B & Weights & \href{https://ai.google.dev/gemma}{https://ai.google.dev/gemma} \\
Ministral-8B-it & Mistral & General & 8B & Weights & \href{https://mistral.ai}{https://mistral.ai} \\
\midrule
ChemDFM-v1.5-8B & SJTU & Chemistry & 8B & Weights & \href{https://github.com/OpenDFM/ChemDFM}{https://github.com/OpenDFM/ChemDFM} \\
SciGLM-6B & Tsinghua & Science & 6B & Weights & \href{https://github.com/THUDM/SciGLM}{https://github.com/THUDM/SciGLM} \\
LlaSMol-Mistral-7B & OSU & Chemistry & 7B & Weights & \href{https://huggingface.co/osunlp/LlaSMol-Mistral-7B}{https://huggingface.co/osunlp/LlaSMol-Mistral-7B} \\
ChemLLM-7B-chat & ShanghaiAILab & Chemistry & 7B & Weights & \href{https://huggingface.co/AI4Chem/ChemLLM-7B-Chat}{https://huggingface.co/AI4Chem/ChemLLM-7B-Chat} \\
\bottomrule
\end{tabular}%
}

\label{tab:model_description}
\end{table*}

\section{Case Studies}\label{ap:case_study}
In this section, we provide several typical bad cases by LLMs.

\begin{tcolorbox}[colback=gray!5, colframe=black, arc=2mm,left=3pt,right=3pt,
    boxsep=5pt,    boxrule=0.5pt,   breakable]  
    \textbf{Ability:} Relevant Information Identification\\
    \\
    \textbf{Question:} \\ 
    Could you determine the chemical formula for the compound identified as mp-775760?\\

    \textbf{Correct Answer:}\\
    "LiFeF3"\\
    
\textbf{Prediction of GPT-4o-mini:}\\
"C17H20ClN3O2S" \quad {\textcolor{red}{×}}

\vspace{0.5em}

\textbf{Prediction of GPT-4o:}\\
"LiFeF3" \quad {\textcolor[RGB]{0,150,0}{$\checkmark$}}

\end{tcolorbox}
\textbf{Remarks:} GPT-4o-mini accurately identified the target column and provided a chemical formula as the response; however, it incorrectly identified the context data row, leading to a mismatch between the generated formula and the corresponding Material ID.

\begin{tcolorbox}[colback=gray!5, colframe=black, arc=2mm,left=3pt,right=3pt,
    boxsep=5pt,    boxrule=0.5pt,   breakable]  
    \textbf{Ability:} Relevant Information Identification\\
    \\
    \textbf{Question:} \\ 
    How is 'infanrix dtap ipv hep b' (Chemical) connected to 'hepatitis b virus infection' (Disease)?\\

    \textbf{Correct Answer:}\\
Chemical-Disease\\
    
\textbf{Prediction of Llama-Maverick:}\\
C \quad {\textcolor{red}{×}}

\vspace{0.5em}

\textbf{Prediction of DeepSeek-V3:}\\
Chemical-Disease\\\quad {\textcolor[RGB]{0,150,0}{$\checkmark$}}

\end{tcolorbox}
\textbf{Remarks:} Llama4-Maverick failed to correctly identify the corresponding relation in the knowledge graph and provided a completely irrelevant answer ("C"), whereas DeepSeek-V3 responded correctly.

\begin{tcolorbox}[colback=gray!5, colframe=black, arc=2mm,left=3pt,right=3pt,
    boxsep=5pt,    boxrule=0.5pt,    breakable]  
    \textbf{Ability:} Information-absence Detection \\
    \\
    \textbf{Question:} \\ 
    Can you enumerate all the PMIDs related to the interaction between id: 25840 and id: 1528? \\

    \textbf{Correct Answer:}\\
    "I cannot answer the question due to insufficient information in the retrieved data."\\
    
    \textbf{Prediction of Claude-3.5-Sonnet:}\\
    "16239215, 15604093" \quad \textcolor{red}{×}
\end{tcolorbox}
\textbf{Remarks:} Claude-3.5-Sonnet failed to detect the absence of question-relevant context in context. Instead, it identified an incorrect Context Row in KG as the relevant context, and thus did not refuse to answer the question, but rather provided an incorrect answer.

\begin{tcolorbox}[colback=gray!5, colframe=black, arc=2mm, left=3pt,right=3pt,
    boxsep=5pt,    boxrule=0.5pt,   breakable]  
    \textbf{Ability:} Multi-source Information Integration\\
    \\
    \textbf{Question:} \\ 
    What are all the pairs of entity names that have a Gene-Gene relationship type?\\

    \textbf{Correct Answer:}\\
    "cyp4f2,ggcx", "hras,kdr", "cyb5r3,cyb5a"\\
    
    \textbf{Prediction of SciGLM-6B:}\\
    "Gene", "Gene"\quad \textcolor{red}{×}
\end{tcolorbox}
\textbf{Remarks:} SciGLM-6B failed to provide the correct answer and merely repeated the vocabulary from the question. It also failed to output the response in the required format.

\begin{tcolorbox}[colback=gray!5, colframe=black, arc=2mm, left=3pt,right=3pt,
    boxsep=5pt,    boxrule=0.5pt,   breakable]  
    \textbf{Ability:} Multi-source Information Integration\\
    \\
    \textbf{Question:} \\ 
    Among the molecules with cid: 138031, 91721881, 131783619, and 104741, which one possesses the highest heavycnt?\\

    \textbf{Correct Answer:}\\
    131783619\\
    
    \textbf{Prediction of ChemLLM-7B-Chat:}\\
    49,36 That\\n281\\n1,64,0585\quad {\textcolor{red}{×}}\\

    \textbf{Prediction of Qwen2.5-7B-it:}\\
    131783619\quad  {\textcolor[RGB]{0,150,0}{$\checkmark$}}\\

\end{tcolorbox}
\textbf{Remarks:} The result from ChemLLM-7B-Chat is entirely unrelated to the question. For LLMs with weaker context understanding capabilities and instruction-following abilities, the occurrence of such responses is a key reason for their poor performance.

\begin{tcolorbox}[colback=gray!5, colframe=black, arc=2mm,left=3pt,right=3pt,
    boxsep=5pt,    boxrule=0.5pt,    breakable]  
    \textbf{Ability:} Context-aware Inference\\
    \\
    \textbf{Question:} \\ 
    Based on the findings of the study, how do ovarian hormones in females influence the metabolic reprogramming effects of Schistosoma mansoni infection on bone marrow-derived macrophages (BMDM)?\\

    \textbf{Options:}\\
    A.Ovarian hormones enhance the metabolic reprogramming...\\
    B.Ovarian hormones do not affect the metabolic...\\
    C.Ovarian hormones inhibit the metabolic reprogramming...\\
    D.Ovarian hormones cause an increase in glycolysis...\\
    \\
    \textbf{Correct Answer:}\\
    C\\
    
    \textbf{Prediction of ChemDFM-v1.5-8B:}\\
    None\\

    \textbf{Error:}\\
    This model's maximum context length is 8192 tokens. However, you requested 13432 tokens in the messages, Please reduce the length of the messages.
\end{tcolorbox}
\textbf{Remarks:} Some individual papers exceed the maximum length limit of certain models. In such cases, we can only classify them as errors.

\begin{tcolorbox}[colback=gray!5, colframe=black, arc=2mm, left=3pt,right=3pt,
    boxsep=5pt,    boxrule=0.5pt,   breakable]  
    \textbf{Ability:} Context-aware Inference\\
    \\
    \textbf{Question:} \\ 
    Based on the first part of the article, what conclusions can be inferred about the role of surface imperfections in the anisotropic Rashba effect observed in the 2D Janus XA2Y monolayers, and what implications might this have for spintronic applications?\\
    \textbf{Options:}\\
    ...\\

    \textbf{Correct Answer:}\\
    A, B\\
    
    \textbf{Prediction of GPT-4o:}\\
    A\quad \textcolor{red}{×}\\
\end{tcolorbox}
\textbf{Remarks:} GPT-4o demonstrates some reasoning ability and selected a correct answer; however, it failed to identify all the correct answers in a multiple-choice question.

\begin{tcolorbox}[colback=gray!5, colframe=black, arc=2mm, left=3pt,right=3pt,
    boxsep=5pt,    boxrule=0.5pt,   breakable] 
\textbf{Ability:} Context-aware Inference\\

\textbf{Question:} \\ 
List all the material IDs that have Space Group Symbol as ``Amm2''.

\textbf{Correct Answer:}\\
mp-1017367, mp-1021395, mp-1219958\\

\textbf{Prediction of Qwen3-8B with thinking:}\\
<think>Okay, so I need to...</think>\\
mp-1219958,mp-1021395,mp-1017367\quad\textcolor[RGB]{0,150,0}{$\checkmark$}\\ 
\vspace{1em}

\textbf{Prediction of Qwen3-8B without thinking:}\\
mp-1205400,mp-1219471,mp-1219958,mp-1221742,mp-1222109,mp-1233960,mp-1245579,mp-1272454,mp-1372845,mp-1406912,mp-14107,mp-1411625,mp-1517069,mp-1518293,mp-1518633,mp-15644,mp-1638589,mp-17955,mp-18026,mp-18027,mp-18028,mp-18029,mp-18030,mp-18031,mp-18032,mp-18033,mp-18034,mp-18035,mp-18036,... \textcolor{red}{×}\\

\textbf{Prediction of Qwen2.5-7B-it:}\\
mp-1219958 \textcolor{red}{×}\\

\end{tcolorbox}
\textbf{Remarks:} The Qwen3-8B model with thinking activation achieved 100\% precision and recall by accurately identifying all three correct material IDs. In contrast, the non-thinking variant generated numerous false positives, indicating a substantial decline in performance. The Chain-of-Thought (CoT) mechanism effectively directed the model’s attention to relevant fields—specifically, by constraining the search explicitly to the "Space Group Symbol" column—thereby preventing the inclusion of erroneous data from unrelated columns or positional artifacts. Moreover, CoT activation suppressed redundant and irrelevant output patterns.

\begin{tcolorbox}[colback=gray!5, colframe=black, arc=2mm, left=3pt,right=3pt,
    boxsep=5pt,    boxrule=0.5pt,   breakable] 
\textbf{Ability:} Context-aware Inference\\

\textbf{Question:} \\ 
Based on the methods and experimental results of the article, which factors could likely contribute to the more effective anti-PD1 response in the orthotopic murine model of mismatch repair-deficient lung adenocarcinoma compared to the subcutaneous model?\\

A: The subclonality of tumor cells resulting in increased clonal neoantigen frequency in the orthotopic model.\\
B: Higher overall immune cell infiltration, particularly T cells and dendritic cells, in the orthotopic lung environment.\\
C: A similar proportion of tumor-associated macrophages and CD8+ T cells but different responses to anti-PD1 treatment in the subcutaneous model.\\
D: Lower mutational burden resulting from fewer unique variants in the orthotopic model, affecting the immune response.\\

\textbf{Correct Answer:}\\
A\\

\textbf{Prediction of DeepSeek-V3:}\\
B \textcolor{red}{×}\\

\textbf{Prediction of DeepSeek-R1:}\\
<think>Alright, let's tackle...</think>\\
A{\textcolor[RGB]{0,150,0}{$\checkmark$}}

\end{tcolorbox}
\textbf{Remarks:} In the task of context-aware inference within unstructured text, DeepSeek-R1 demonstrates stronger reasoning capabilities than DeepSeek-V3. This indicates that reasoning models also possess significant advantages in understanding unstructured text.

\begin{tcolorbox}[colback=gray!5, colframe=black, arc=2mm, left=3pt,right=3pt,
    boxsep=5pt,    boxrule=0.5pt,   breakable] 
\textbf{Ability:} Multi-source Information Integration\\

\textbf{Question:} \\ 
What entity engages in interactions with the entities: epilepsy and diabetes mellitus?\\

\textbf{Correct Answer:}\\
gabapentin\\

\textbf{Prediction of DeepSeek-R1:}\\
<think>Okay, let's...</think>\\
No entity in the provided data interacts with both epilepsy and diabetes mellitus.\quad \textcolor{red}{×}\\

\end{tcolorbox}
\textbf{Remarks:} Even the well-performing reasoning model DeepSeek-R1 struggles to answer complex questions involving knowledge graph multi-source information integration.

\end{document}